\documentclass[10pt,twocolumn,letterpaper]{article}

\usepackage{cvpr}              

\usepackage[dvipsnames]{xcolor}

\definecolor{cvprblue}{rgb}{0.21,0.49,0.74}
\usepackage[pagebackref,breaklinks,colorlinks,citecolor=cvprblue]{hyperref}

\def\paperID{12665} 
\def\confName{CVPR}
\def\confYear{2024}

\title{Noisy-Correspondence Learning for Text-to-Image Person Re-identification}

\usepackage{algorithm}
\usepackage{algorithmic}
\usepackage{multirow}
\usepackage{array}  

\usepackage{pifont}

\definecolor{mydeepgreen}{RGB}{0,100,0} 

\newtheorem{lemma}{Lemma}
\newtheorem{proof}{Proof}

\author{Yang Qin\textsuperscript{\rm 1}\and Yingke Chen\textsuperscript{\rm 2}\and Dezhong Peng\textsuperscript{\rm 1}\and Xi Peng\textsuperscript{\rm 1}\and Joey Tianyi Zhou\textsuperscript{\rm 3}\and Peng Hu\textsuperscript{\rm 1,*}
\\
College of Computer Science, Sichuan University\\
Department of Computer and Information Sciences, Northumbria University\\
\and 
Centre for Frontier AI Research (CFAR) and Institute of High Performance Computing (IHPC), A*STAR, Singapore 
}

\author{
    Yang Qin\textsuperscript{\rm 1}\quad Yingke Chen\textsuperscript{\rm 2}\quad Dezhong Peng\textsuperscript{\rm 1,4,5}\quad Xi Peng\textsuperscript{\rm 1}\quad Joey Tianyi Zhou\textsuperscript{\rm 3}\quad Peng Hu\textsuperscript{\rm 1}\thanks{Corresponding author: Peng Hu (penghu.ml@gmail.com).} \\
    \textsuperscript{\rm 1}{\normalsize College of Computer Science, Sichuan University, Chengdu, 610095,  China.} 
    \\\textsuperscript{\rm 2}{\normalsize Department of Computer and Information Sciences, Northumbria University, Newcastle upon Tyne NE1 8ST, UK.} 
    \\  \textsuperscript{\rm 3}{\normalsize Centre for Frontier AI Research (CFAR) and Institute of High Performance Computing (IHPC), A*STAR, Singapore.}
     \\  \textsuperscript{\rm 4}{\normalsize Sichuan Newstrong UHD Video Technology Co., Ltd., Chengdu 610095, China.}
      \\  \textsuperscript{\rm 5}{\normalsize Chengdu Ruibei Yingte Information Technology Company Ltd., Chengdu 610065, China.}
}

\begin{document}
\maketitle

\begin{abstract}
    Text-to-image person re-identification~(TIReID) is a compelling topic in the cross-modal community, which aims to retrieve the target person based on a textual query. Although numerous TIReID methods have been proposed and achieved promising performance, they implicitly assume the training image-text pairs are correctly aligned, which is not always the case in real-world scenarios. In practice, the image-text pairs inevitably exist under-correlated or even false-correlated, \textit{a.k.a} noisy correspondence (NC), due to the low quality of the images and annotation errors. To address this problem, we propose a novel Robust Dual Embedding method (RDE) that can learn robust visual-semantic associations even with NC. Specifically, RDE consists of two main components: 1) A Confident Consensus Division (CCD) module that leverages the dual-grained decisions of dual embedding modules to obtain a consensus set of clean training data, which enables the model to learn correct and reliable visual-semantic associations. 2) A Triplet Alignment Loss (TAL) relaxes the conventional Triplet Ranking loss with the hardest negative samples to a log-exponential upper bound over all negative ones, thus preventing the model collapse under NC and can also focus on hard-negative samples for promising performance. We conduct extensive experiments on three public benchmarks, namely CUHK-PEDES, ICFG-PEDES, and RSTPReID, to evaluate the performance and robustness of our RDE. Our method achieves state-of-the-art results both with and without synthetic noisy correspondences on all three datasets.  Code is available at \href{https://github.com/QinYang79/RDE}{https://github.com/QinYang79/RDE}.
\end{abstract}

\section{Introduction}
\label{sec:intro}

Text-to-image person re-identification (TIReID)~\cite{li2017person,shu2022see,jiang2023cross} aims to understand the natural language descriptions to retrieve the matched person image from a large gallery set. This task has received increasing attention from both academic and industrial communities recently, \eg, finding/tracking suspect/lost persons in a surveillance system~\cite{wang2015multi,eom2019learning}. However, TIReID remains a challenging task due to the inherent heterogeneity gap across different modalities and appearance attribute redundancy.

\begin{figure}[t]
    \centering
    \resizebox{\linewidth}{!}{ 
    \includegraphics{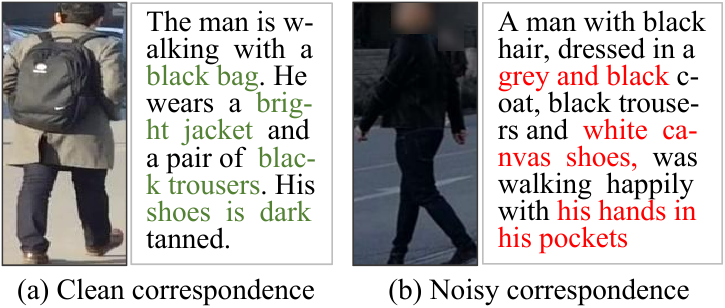}}
    \caption{
    The illustration of noisy correspondence. The figure shows an example of the NC problem, which occurs when the image-text pairs are wrongly aligned, \ie, false positive pairs (FPPs). Since the model does not know which pairs are noisy in practice, they will unavoidably degrade the performance by incorrect supervision information. As seen in the figure, (a) the clean image-text pair is semantically matched, while (b) the noisy pair is not, which would cause the cross-modal model to learn erroneous visual-textual associations. Note that both examples in (a) and (b) are from and actually exist in the RSTPReid dataset~\cite{zhu2021dssl}.}
    \label{fig0}
\vspace{-0.3cm}
\end{figure}

To tackle these challenges, most of the existing methods explore global- and local-matching alignment to learn accurate similarity measurements for person re-identification. To be specific, some global-matching methods~\cite{zhang2018deep,wu2021lapscore,shu2022see} leverage vision/language backbones to extract modality-specific features and employ contrastive learning to achieve global visual-semantic alignments. To capture fine-grained information, some local-matching methods~\cite{niu2020improving,jing2020pose,wang2021text,shao2022learning} explicitly align local body regions to textually described entities/objectives to improve the discriminability of pedestrian features. Recently, some works~\cite{han2021text,yan2022clip,jiang2023cross} propose to exploit visual/semantic knowledge learned by the pre-trained models, such as BERT~\cite{devlin2018bert}, ViT~\cite{dosovitskiy2020image}, and CLIP~\cite{radford2021learning}, and achieve explicit global alignments or discover more fine-grained local correspondence, thus boosting the re-identification performance. Although these methods achieve remarkable progress, they implicitly assume that all training image-text pairs are aligned correctly.

In reality, this assumption is hard or even impossible to hold due to the person's pose, camera angle, illumination, and other inevitable factors in images, which may result in some inaccurate/mismatched textual descriptions of images (see \Cref{fig0}), \eg, the RSTPReid dataset~\cite{zhu2021dssl}. Moreover, we observe that excessive such imperfect/mismatched image-text pairs would cause an overfitting problem and degrade the performance of existing TIReID methods shown in~\Cref{fig4}. Based on the observation, in this paper, we reveal and study a new problem in TIReID, \ie, noisy correspondence (NC). Different from noisy labels, NC refers to the false correspondences of image-text pairs in TIReID, \ie, False Positive Pairs (FPPs): some negative image-text pairs are used as positive ones for cross-modal learning. Inevitably, FPPs will misguide models to overfit noisy supervision and collapse to suboptimal solutions due to the memorization effect~\cite{arpit2017closer} of Deep Neural Networks (DNNs).

To address the NC problem, we propose a Robust Dual Embedding method (RDE) for TIReID in this paper, which benefits from an effective Confident Consensus Division mechanism (CCD) and a novel Triplet Alignment Loss (TAL). Specifically, CCD fuses the dual-grained decisions to consensually divide the training data into clean and noisy sets, thus providing more reliable correspondences for robust learning. To diversify the model grain, the basic global embeddings (BGE) and token selection embeddings (TSE) are presented for coarse-grained and fine-grained cross-modal interactions respectively, thus capturing visual-semantic associations comprehensively.  Different from the widely-used Triplet Ranking loss with the hardest negatives, our TAL relaxes the similarity learning from the hardest negative samples to all negative ones by applying an upper bound, which brings a stable solution for the collapse of training under NC while also benefiting from the hardest negatives mining to achieve promising performance. As a result, our RDE can achieve robustness against NC thanks to the proposed reliable supervision and stable triplet loss. The contributions and innovations of this paper are summarized as follows:

\begin{itemize}
    \item We reveal and study a new and ubiquitous problem in TIReID, termed noisy correspondence (NC). Different from class-level noisy labels, NC refers to erroneous correspondences in the person-description pairs that can mislead the model to learn incorrect visual-semantic associations. To the best of our knowledge, this paper could be the first work to explore this problem in TIReID.
    \item We propose a robust method, termed RDE, to mitigate the adverse impact of NC through the proposed Confident Consensus Division (CCD) and novel Triplet Alignment Loss (TAL). By using CCD and TAL, RDE can obtain convincing consensus pairs and reduce the misleading risks in training, thus embracing robustness against NC.
    \item Extensive experiments on three public image-text person benchmarks demonstrate the robustness and superiority of our method. Our method achieves the best performance both with and without synthetic noisy correspondence on all three datasets.
\end{itemize}

\section{Related Work}
\subsection{Text-to-Image Person Re-identification}
Text-to-image person re-identification (TIReID) is a novel and challenging task that aims to match a person image with a given natural language description~\cite{li2017person,zhang2018deep,bai2023rasa,yan2023learning,shen2023pedestrian,ma2023beat,bai2023text,cao2023empirical,shao2023unified,li2023dcel}. Existing TIReID methods could be roughly classified into two groups according to their alignment levels, \ie, \textbf{global-matching} methods~\cite{shu2022see,zheng2020dual,zhu2021dssl} and \textbf{local-matching} methods~\cite{gao2021contextual,wang2021text,shao2022learning}. The former try to learn cross-modal embeddings in a common latent space by employing textual and visual backbones with a matching loss (\eg, CMPM/C loss~\cite{zhang2018deep} and Triplet Ranking loss~\cite{faghri2017vse++}) for TIReID. However, these methods mainly focus on global features while ignoring the fine-grained interactions between local features, which limits their performance improvement. To achieve fine-grained interactions, some of the latter methods explore explicit local alignments between body regions and textual entities for more refined alignments. However, these methods require more computational resources due to the complex local-level associations. Recently, inspired and benefited from \textbf{vision-language pre-training} models~\cite{radford2021learning}, some methods~\cite{han2021text,yan2022clip,jiang2023cross} expect to use the learned rich alignment knowledge of pre-trained models for local- or global-alignments. Although these methods achieve promising performance, almost all of them implicitly assume that all input training pairs are correctly aligned, which is hard to meet in practice due to the ubiquitous noise. In this paper, we address the inevitable and challenging noisy correspondence problem in TIReID.

\begin{figure*}[h]
    \centering
    \resizebox{0.8\linewidth}{!}{ 
    
    \includegraphics{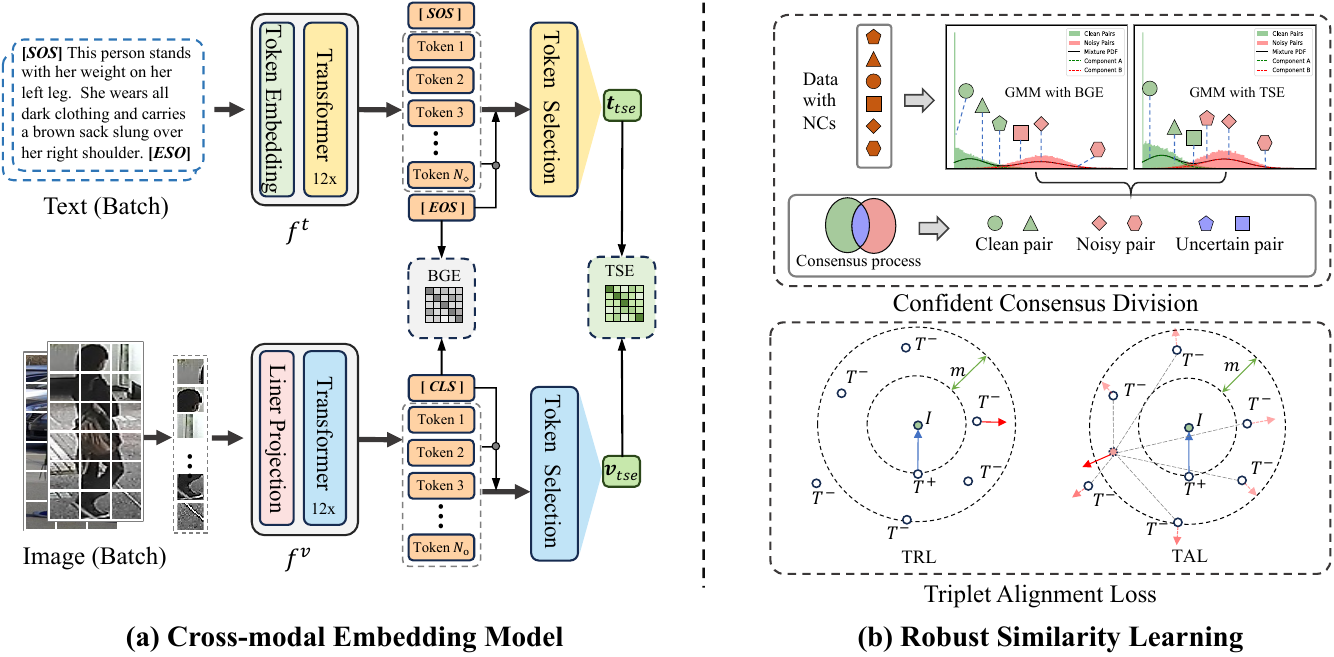}}
    \caption{The overview of our RDE. (a) is the illustration of the cross-modal embedding model used in RDE, which consists of \emph{basical global embedding} (BGE) and \emph{token selection embedding} (TSE) modules with different granularity. By integrating them, RDE can capture coarse-grained cross-modal interactions while selecting informative local token features to encode more fine-grained representations for a more accurate similarity. (b) shows the core of RDE to achieve robust similarity learning, which consists of Confident Consensus Division (CCD) and Triplet Alignment Loss (TAL). CCD performs consensus division to obtain confident clean training data, thus avoiding misleading from noisy pairs. Unlike traditional Triplet Ranking Loss (TRL)~\cite{faghri2017vse++}, TAL exploits an upper bound to consider all negative pairs, thus embracing more stable learning.}
    \label{fig1}
    \vspace{-0.3cm}
\end{figure*}

\subsection{Learning with Noisy Correspondence}
As a special learning paradigm with noisy labels~\cite{li2020dividemix,lu2022ensemble,Feng_2023_CVPR} in multi-modal/view community~\cite{hu2021learning,qin2023edmc,qin2023elastic,qin2023edmc,yang2022robust}, the studies for noisy correspondence (NC) have recently attracted more and more attention in various tasks, \eg, video-text retrieval~\cite{zhang2023robust}, visible-infrared person re-identification~\cite {yang2022learning,yang2024robust,lin2024multi}, and image-text matching~\cite{huang2021learning,qin2022deep}, which means that the negative pairs are wrongly treated as positive ones, \ie, false positive pairs (FPPs). To handle this problem, numerous methods are proposed to learn with NC, which can be broadly categorized into \textbf{sample selection}~\cite{huang2021learning,zhang2023robust,han2023noisy} and \textbf{robust loss functions}~\cite{yang2022learning,qin2022deep,hu2023cross,qin2023cross}. The former commonly leverage the memorization effect of DNNs~\cite{arpit2017closer} to gradually distinguish the noisy data, thus paying more attention to clean data while less attention to noisy data. Differently, the latter methods aim to develop noise-tolerance loss functions to improve the robustness of model training against NC. Although the aforementioned methods achieve promising performance in various tasks, they are not specifically designed for TIReID and may be inefficient or ineffective in person re-identification. In this paper, we propose a well-designed method to tackle the NC problem in TIReID, which not only performs superior in noisy scenarios but also achieves promising performance in ordinary scenarios.

\section{Methodology}
\subsection{Problem Statement}
The purpose of TIReID is to retrieve a pedestrian image from the gallery set that matches the given textual description. For clarity, we represent the gallery set as $\mathcal{V}=\{I_i,y^p_i,y^v_i\}^{N_v}_{i=1}$ and the corresponding text set as $\mathcal{T}=\{T_i,y^v_i\}^{N_t}_{i=1}$, where $N_v$ is the number of images, $N_t$ is the number of texts, $y^p_i\in\mathcal{Y}_p=\{1,\cdots,C\}$ is the class label (person identify), $C$ is the number of identifies, and $y^v_i\in\mathcal{Y}_v=\{1,\cdots,N_v\}$ is the image label. The image-text pair set used in TIPeID can be defined as $\mathcal{P}=\{(I_i,T_i), y^v_i, y^p_i\}^{N}_{i=1}$, where the cross-modal samples of each pair have the same image label $y_i^v$ and class label $y_i^p$. We define a binary correspondence label $l_{ij}\in\{0,1\}$ to indicate the matched degree of any image-text pair. If $l_{ij}=1$, the pair $(I_i,T_j)$ is matched (positive pair), otherwise it is not (negative pair). In practice, due to ubiquitous annotation noise, some unmatched pairs $(l_{ij}=0)$ are wrongly labeled as matched $(l_{ij}=1)$, resulting in noisy correspondences (NCs) and performance degradation. To handle NC for robust TIReID, we present an RDE that leverages the Confident Consensus Division (CCD) and Triplet Alignment Loss (TAL) to mitigate the negative impact of label noise.

\subsection{Cross-modal Embedding Model}
In this section, we describe the cross-modal model used in our RDE. Following previous work~\cite{jiang2023cross}, we utilize the visual encoder $f^v$ and textual encoder $f^t$ of the pre-trained model CLIP as modality-specific encoders to obtain token representations and implement cross-modal interactions through two embedding modules.

\subsubsection{Token Representations} Give an input image $I_i\in \mathcal{V}$, we use the visual encoder $f^v$ of CLIP to tokenize the image into a discrete token representation sequence with a length of $N_{\circ}+1$, \ie, $\boldsymbol{V}_{i}=f^v(I_i)=\{ \boldsymbol{v}^{i}_g,\boldsymbol{v}^{i}_1,\boldsymbol{v}^{i}_2,\cdots, \boldsymbol{v}^{i}_{N_\circ} \}^\top\in\mathbb{R}^{(N_{\circ}+1)\times d}$, where $d$ is the dimensionality of the shared latent space. These features include an encoded feature $\boldsymbol{v}^{i}_g$ of the [\emph{CLS}] token and patch-level local features $\{\boldsymbol{v}^{i}_j\}^{N_{\circ}}_{j=1}$ of $N_{\circ}$ fixed-sized non-overlapping patches of $I_i$, wherein $\boldsymbol{v}^{i}_g$ can represent the global representation. For an input text $T_i\in\mathcal{T}$, we apply the textual encoder $f^t$ of CLIP to obtain global and local representations. Specifically, following IRRA~\cite{jiang2023cross}, we first tokenize the input text $T_i$ using lower-cased byte pair encoding (BPE) with a 49,152 vocab size into a token sequence. The token sequence is bracketed with [\emph{SOS}] and [\emph{EOS}] tokens to represent the beginning and end of the sequence. Then, we feed the token sequence into $f_t$ to obtain the features $\boldsymbol{T}_i=\{ \boldsymbol{t}^{i}_s,\boldsymbol{t}^{i}_1,\cdots,\boldsymbol{t}^{i}_{N_\diamond},\boldsymbol{t}^{i}_e \}^\top\in\mathbb{R}^{(N_{\diamond}+2)\times d}$, where $\boldsymbol{t}^{i}_s$ and $\boldsymbol{t}^{i}_e$ are the features of [\emph{SOS}] and [\emph{EOS}] tokens and $\{\boldsymbol{v}_j^i\}^{N_\diamond}_{j=1}$ are the word-level local features of $N_\diamond$ word tokens of text $T_i$. Generally, the $\boldsymbol{t}^{i}_e$ can be regarded as the sentence-level global feature of $T_i$.

\subsubsection{Dual Embedding Modules} To measure the similarity between any image-text pair $(I_i,T_j)$, we can directly use the global features of [\emph{CLS}] and [\emph{EOS}] tokens to compute the \emph{Basic Global Embedding} (BGE) similarity by the cosine similarity, \ie, $S^b_{ij} = {{\boldsymbol{v}^i_g}{\boldsymbol{t}^j_e}^\top} \big/{\|\boldsymbol{v}^i_g\|\|\boldsymbol{t}^j_e\|}$, where the global features represent the global embedding representations of two modalities. However, optimizing the BGE similarities alone may not capture the fine-grained interactions between two modalities, which will limit performance improvement. To address this issue, we exploit the local features of informative tokens to learn more discriminative embedding representations, thus mining the fine-grained correspondences. In CLIP, the global features of the tokens ([\emph{CLS}] and [\emph{EOS}]) are obtained by a weighted aggregation of all local token features. These weights reflect the correlation between the global token and each local token. Following previous methods~\cite{yan2022clip,zhu2022dual}, we could select the informative tokens based on these correlation weights to aggregate local features for a more representative global embedding.

In practice, these correlation weights can be obtained directly in the self-attention map of the last Transformer blocks of $f^v$ and $f^t$, which reflects the relevance among the input $1+N_\circ$ (or $2+N_\diamond$) tokens. Given the output self-attention map $\boldsymbol{A}^v_i\in\mathbb{R}^{(1+N_\circ)\times(1+N_\circ)}$ of image $I_i$, the correlation weights between global token and local tokens are $\{a^v_{i,j}\}^{N_\circ}_{j=1}=\boldsymbol{a}^v_i=\boldsymbol{A}^v_i[0,1:N_\circ+1 ]\in\mathbb{R}^{N_\circ}$.
Similarly, for text $T_i$, the correlation weights are $\{a^t_{i,j}\}^{N_\diamond}_{j=1}=\boldsymbol{a}^t_i=\boldsymbol{A}^t_i[0,1:N_\diamond+1]\in\mathbb{R}^{N_\diamond}$, where $\boldsymbol{A}^t_i\in\mathbb{R}^{(2+N_\diamond)\times(2+N_\diamond)}$ is the output self-attention map for text $I_i$. 
Then, we select a proportion ($TopK$) of the corresponding token features with higher scores for embedding. Specifically, for $I_i$, the selected token sequences and correlation weights are reorganized as $\boldsymbol{V}^s_i=\{ \boldsymbol{v}^i_j \}_{j\in\boldsymbol{K}^v_i }$
and $\hat{\boldsymbol{a}}^v_i=\{ a^v_{i,j}\}_{j\in\boldsymbol{K}^v_i }$, where $\boldsymbol{K}^v_i\in \mathbb{R}^{\lfloor \mathcal{R} N_\circ \rfloor}$ is the set of indices for the selected local tokens of $I_i$ and $\mathcal{R}$ is the selection ratio. For text $T_i$, the selected token sequences and correlation weights are also reorganized as $\boldsymbol{T}^s_i=\{ \boldsymbol{t}^i_j \}_{j\in\boldsymbol{K}^t_i }$
and $\hat{\boldsymbol{a}}^t_i=\{ a^t_{i,j}\}_{j\in\boldsymbol{K}^t_i }$, where
$\boldsymbol{K}^t_i\in \mathbb{R}^{\min(\lfloor \mathcal{R} N_\diamond^\prime \rfloor, N_\diamond)}$ is the set of indices for the selected local tokens of $T_i$. $N_\diamond^\prime$ is the maximum input sequence length of $f^t$.
For $I_i$ and $T_i$, we perform an embedding transformation on these selected token features to obtain subtle representations, instead of using complex fine-grained correspondence discovery used in CFine~\cite{yan2022clip}. The transformation is performed by an embedding module like the residual block~\cite{he2016deep}, as follows:
\begin{equation}
    \begin{aligned}
        \boldsymbol{v}^i_{tse} =&{MaxPool}({MLP(\hat{\boldsymbol{V}}^s_i)}+FC(\hat{\boldsymbol{V}}^s_i)),\\
        \boldsymbol{t}^i_{tse} =&{MaxPool}({MLP(\hat{\boldsymbol{T}}^s_i)}+FC(\hat{\boldsymbol{T}}^s_i)), 
    \end{aligned}
\end{equation}
where $MaxPool(\cdot)$ is the max-pooling function, $MLP(\cdot)$ is a multi-layer perceptron (MLP) layer, $FC(\cdot)$ is a linear layer, $\hat{\boldsymbol{V}}^s_i=L2Norm({\boldsymbol{V}}^s_i)$, and $\hat{\boldsymbol{T}}^s_i=L2Norm({\boldsymbol{T}}^s_i)$. $L2Norm(\cdot)$ is the $\ell_2$-normalization function to normalize features. 
Finally, for any pair $(I_i,T_j)$, we compute the cosine similarity $S^t_{ij}$ between $\boldsymbol{v}_{tse}^i$ and $ \boldsymbol{t}_{tse}^j$ as the \emph{Token Selection Embedding} (TSE) similarity to measure the cross-modal matching degree for auxiliary training and inference.

\subsection{Robust Similarity Learning}
In this section, we detail how we use the image-text similarities computed by the dual embedding modules for robust TIReID, which involves Confident Consensus Division (CCD) and Triplet Alignment Loss (TAL). 

\subsubsection{Confident Consensus Division}
To alleviate the negative impact of NC, the key is to filter the possible noisy pairs in the training data, which directly avoids false supervision information. Some previous work in learning with noisy labels~\cite{han2018co,li2020dividemix,huang2021learning} are inspired by the memorization effect~\cite{arpit2017closer} of DNNs to perform filtrations, \ie, the clean (easy) data tend to have a smaller loss value than that of noisy (hard) data in early training. Based on this, we can exploit the two-component Gaussian Mixture Model (GMM) to fit the per-sample loss distributions computed by the predictions of BGE and TSE to further identify the noisy pairs in the training data. Specifically, given a cross-modal model $\mathcal{M}$, we first define the per-sample loss as:
\begin{equation}
    \ell(\mathcal{M},\mathcal{P}) = \{\ell_i\}^N_{i=1}= \big\{\mathcal{L}(I_i,T_i)\big\}^N_{i=1},
\end{equation}
where $\mathcal{L}$ is the loss function for pair $(I_i,T_i)\in\mathcal{P}$ to bring them closer in the shared latent space. In our method, $\mathcal{L}$ is the proposed $\mathcal{L}_{tal}$ defined in~\Cref{eqtal}. Then, the per-sample loss is fed into the GMM to separate clean and noisy data, \ie, assigning the Gaussian component with a lower mean value as a clean set and the other as a noisy one, respectively. Following~\cite{li2020dividemix,huang2021learning}, we use the Expectation-Maximization algorithm to optimize the GMM and compute the posterior probability $p(k|\ell_i)=p(k)p(\ell_i|k)/p(\ell_i)$ for the $i$-th pair as the probability of being clean/noisy pair, where $k\in\{0,1\}$ is used to indicate whether it is a clean or a noisy component. 
Then, we set a threshold $\delta=0.5$ to $\{p(k=0|l_i)\}^N_{i=1}$ to divide the data into clean and noisy sets, \ie,
\begin{equation} 
\begin{aligned}
        \mathcal{P}^c=&\{(I_i,T_i)|p(k=0|\ell_i)>\delta,\forall (I_i,T_i)\in\mathcal{P} \},\\
        \mathcal{P}^n=&\{(I_i,T_i)|p(k=0|\ell_i)\leq \delta,\forall (I_i,T_i)\in\mathcal{P} \},
\end{aligned}
\label{eq3}
\end{equation}
where $\mathcal{P}^c$ and $\mathcal{P}^n$ are the divided clean and noisy sets, respectively. For BGE and TSE, the divisions conducted with~\Cref{eq3} are $\mathcal{P} = \mathcal{P}^c_{bge}\cup \mathcal{P}^n_{bge}$ and $\mathcal{P} = \mathcal{P}^c_{tse}\cup \mathcal{P}^n_{tse}$, separately.

To obtain the final reliable divisions, we propose to exploit the consistency between the two divisions to find the consensus part as the final confident clean set, \ie, $\hat{\mathcal{P}}^c=\mathcal{P}^c_{bge}  \cap  \mathcal{P}^c_{tse}$. The rest of the data can be divided into noisy and uncertain subsets, \ie, $\hat{\mathcal{P}}^n=\mathcal{P}^n_{bge}  \cap  \mathcal{P}^n_{tse}$ and $\hat{\mathcal{P}}^u= \mathcal{P} - (\hat{\mathcal{P}}^c\cup  \hat{\mathcal{P}}^n)$. Finally, we exploit the divisions to further recalibrate the correspondence labels, \eg, for $i$-th pair, the process can be expressed as:
\begin{equation}\label{hl_ii}
     \hat{l}_{ii} = \left\{ \begin{array}{ll}
        1,&\text{if $(I_i,T_i) \in \hat{\mathcal{P}}^c$}, \\ 
        0,&\text{if $(I_i,T_i) \in \hat{\mathcal{P}}^n$},\\
        Rand(\{0, 1\}),&\text{if $(I_i,T_i) \in \hat{\mathcal{P}}^u$},
\end{array}\right.
\end{equation}
where $Rand(X)$ is the function to randomly select an element from the collection $X$. 

\subsubsection{Triplet Alignment Loss}
The Triplet Ranking Loss (TRL) is a common matching loss that is widely used in cross-modal learning, and achieves promising performance by employing the hardest negatives, \eg, image-text matching~\cite{diao2021similarity}, video-text retrieval~\cite{dong2021dual}, \etc However, we find that this strategy may lead to bad local minima or even model collapse for TIReID under NC in the early stages of training. In contrast, the summation version of TRL that considers all negative samples, namely TRL-S, can maintain better stability and avoid model collapse, but suffers from insufficient performance due to the lack of attention to hard negatives (see \Cref{revisit} for more discussion). Therefore, we propose a novel Triplet Alignment Loss (TAL) to guide TIReID, which differs from TRL in that it relaxes the optimization of the hardest negatives to all negatives with an upper bound (see \Cref{lam1}). Thanks to the relaxation, TAL reduces the risk of the optimization being dominated by the hardest negatives, thereby making the training more stable and comprehensive by considering all pairs. For an input pair $(I_i,T_i)$ in a mini-batch $\mathbf{x}$, TAL is defined as
\begin{equation}
 \resizebox{\linewidth}{!}{$
    \begin{aligned}
        \mathcal{L}_{tal}(I_i,T_i)&=\big[ m -  S^+_{i2t}(I_i) + \tau \log ( \sum_{j = 1}^K q_{ij}\exp(S(I_i,T_j)/\tau)  ) \big]_+\\ &+\big[ m -  S^+_{t2i}(T_i)  + \tau \log ( \sum_{j = 1}^K q_{ji}\exp(S(I_j,T_i)/\tau)  ) \big]_+,
    \end{aligned}$}
    \label{eqtal}
\end{equation}
where $m$ is a positive margin coefficient, $\tau$ is a temperature coefficient to control hardness, $S(I_i,T_j)\in \{S_{ij}^b, S_{ij}^t\}$, $[x]_+ \equiv \max (x,0)$, $\exp(x)\equiv e^x$, $q_{ij}=1-l_{ij}$,  and $K$ is the size of $\mathbf{x}$. From~\Cref{lam1}, as $\tau\to0$, TAL approaches TRL and focuses more on hard negatives. Since multiple positive pairs from the same identity may appear in the mini-batch, $S^+_{i2t}(I_i) =\sum^K_{j=1} \alpha_{ij}  S(I_i,T_j)$ is the weighted average similarity of positive pairs for image $I_i$, where $\alpha_{ij}=\frac{l_{ij}\exp{(S(I_i,T_j)/\tau)}}{\sum^N_{k=1}l_{ik}\exp{(S(I_i,T_k)/\tau)}}$. Similarly, $S^+_{i2t}(T_i)$ is the weighted average similarity of positive pairs for text $T_i$.
\begin{lemma}
    TAL is the upper bound of TRL, i.e.,
    \begin{equation}
    \begin{aligned}
        &\mathcal{L}_{trl}(I_i,T_i)= \big[ m -  S^+_{i2t}(I_i) +  S(I_i,\hat{T}_i)  \big]_+    \\
&+\big[ m -  S^+_{t2i}(T_i)  + S(\hat{I}_i,T_i) \big]_+ \leq \mathcal{L}_{tal}(I_i,T_i),
    \end{aligned}
    \end{equation}
    where $\hat{T}_i\in\{ T_j|l_{ij}=0, \forall j\in \{1,\cdots,K\} \}$ is the hardest negative text for $I_i$ and $\hat{I}_i\in\{ I_j|l_{ji}=0, \forall j\in \{1,\cdots,K\} \}$ is the hardest negative image for $I_i$, respectively. 
    \label{lam1}
\end{lemma}

\subsubsection{Revisit Triplet Raking Loss}\label{revisit}
\begin{figure}[!h]
\centering
\begin{subfigure}{0.32\linewidth}
\hfill
\includegraphics[width=1\linewidth]{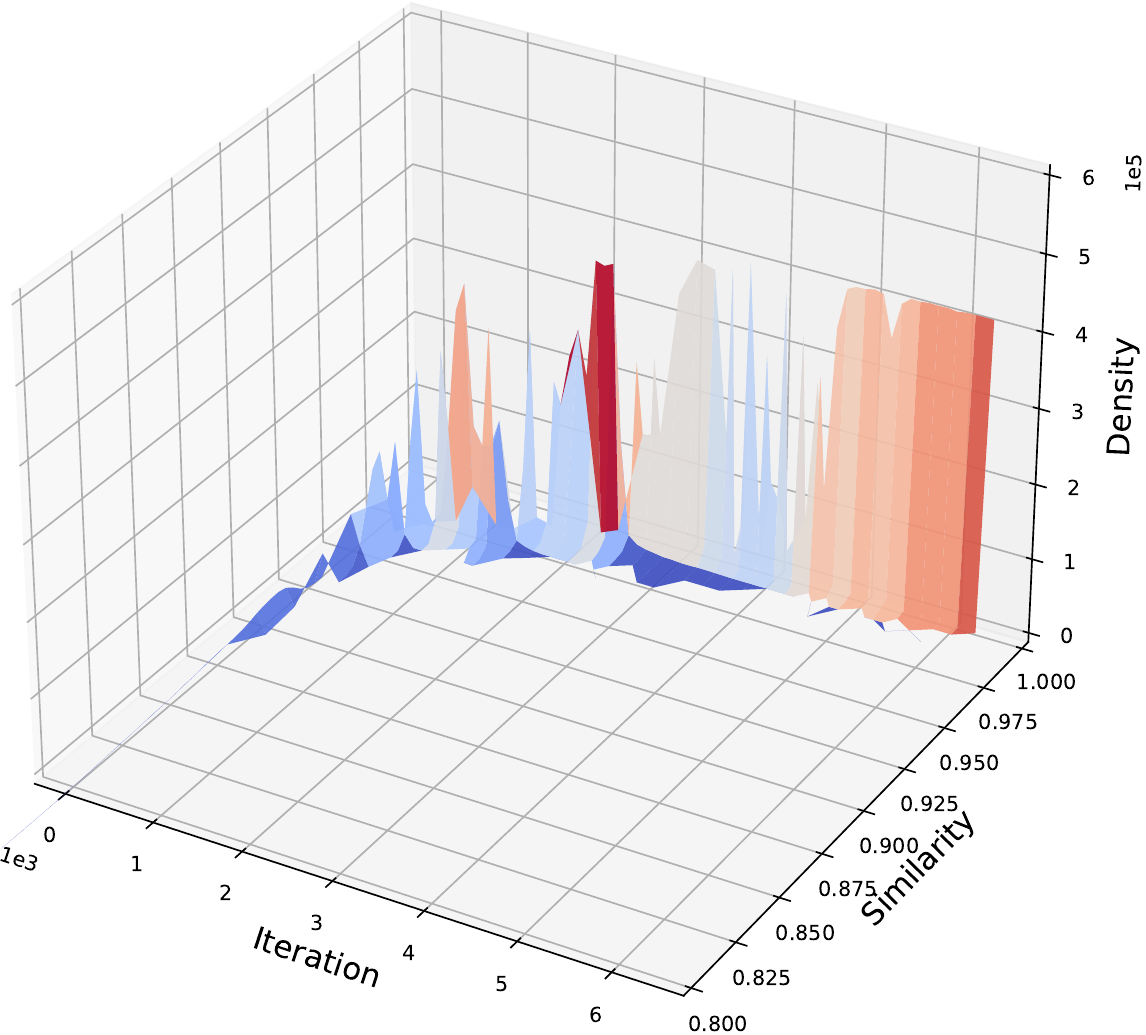}
\centering
\caption{\footnotesize {TRL (3.64\%)}}
\label{trl_3d}
\end{subfigure}
\begin{subfigure}{0.32\linewidth}
\hfill
\includegraphics[width=1\linewidth]{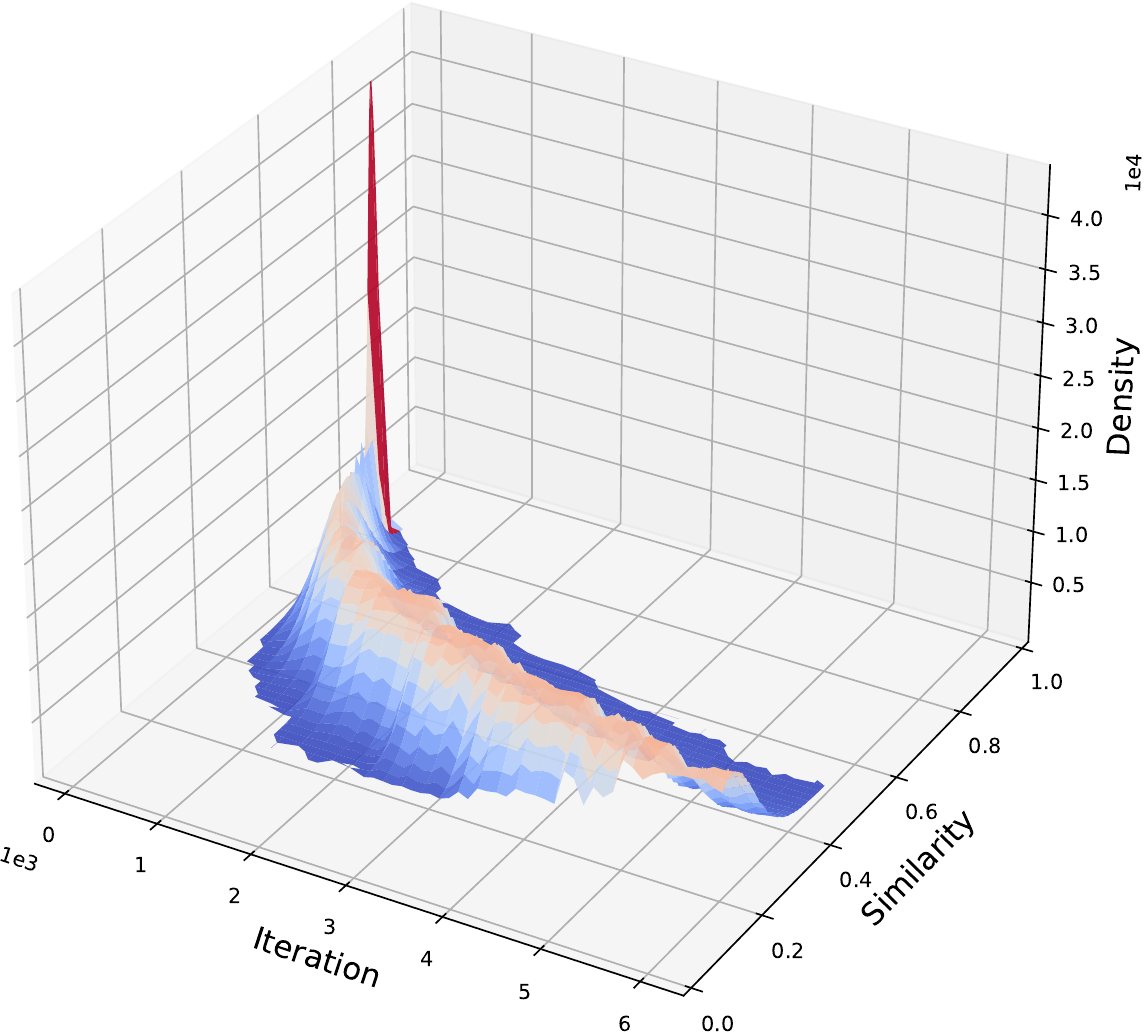}
\centering
\caption{\footnotesize  {TRL-S (44.93\%)}}\label{trls_3d}
\end{subfigure}
\begin{subfigure}{0.32\linewidth}
\hfill
\includegraphics[width=1\linewidth]{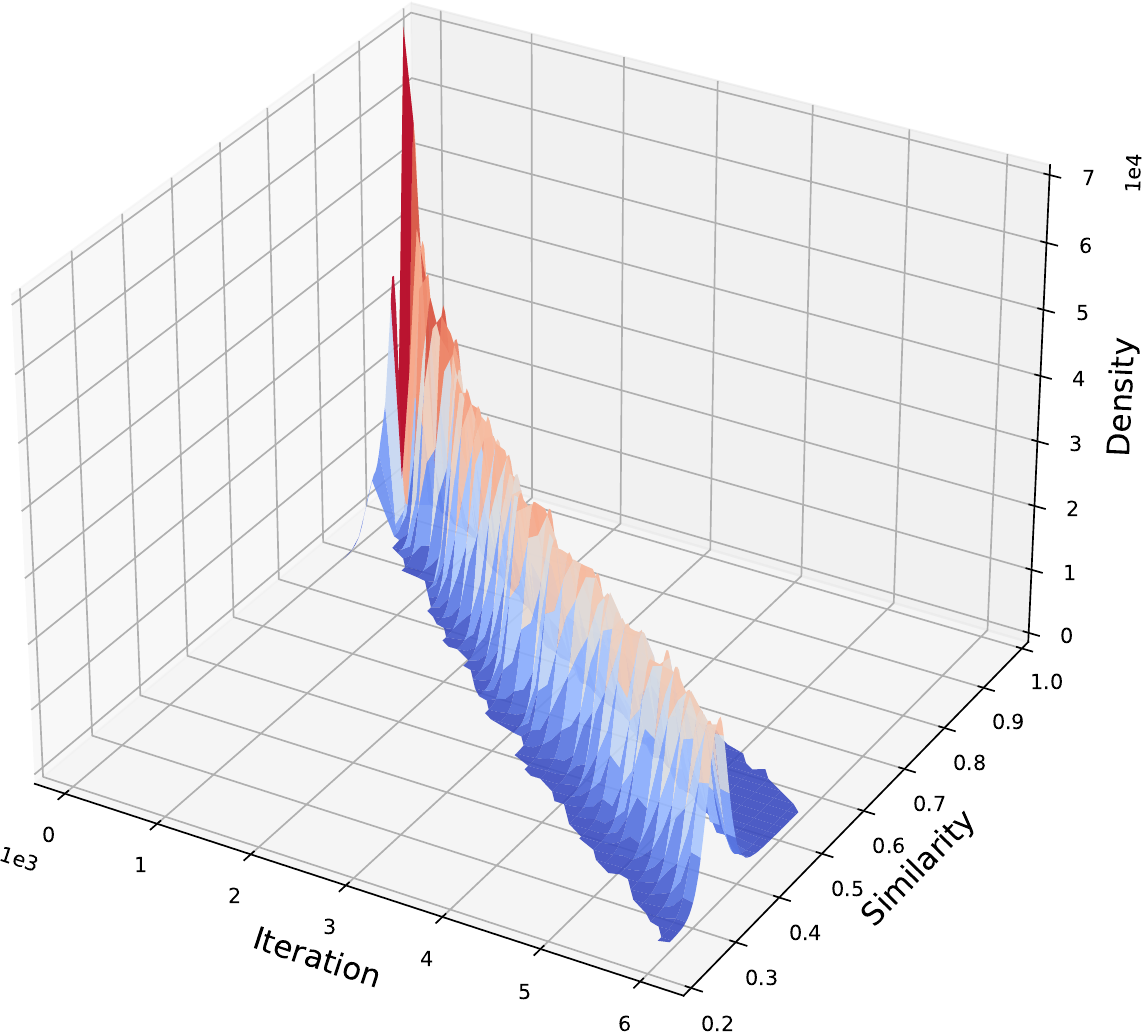}
\centering
\caption{\footnotesize  {TAL (63.35\%)}}
\end{subfigure}
\caption{The difference between TRL, TRL-S, and proposed TAL on the similarity distribution versus iterations. The $y$-$z$ plane represents the similarity density. The corresponding Rank-1 scores of testing are placed in brackets for convenience.
}
\label{fig_sims} 
\end{figure}

To explore the behaviors of the triplet losses in the noisy case, we record the similarity distributions versus iterations of TRL, TRL-S, and the proposed TAL under 50\% noise. From~\Cref{trl_3d}, one can see that the similarities of all pairs are gradually gathered to 1 during training with TRL, \ie, all samples \emph{collapses} to a narrow neighborhood space on a hypersphere, resulting in a trivial solution and a bad performance (3.64\%).
To delve deeper into the underlying reason, we performed a gradient analysis. For ease of representation and analysis, we only consider one direction since image-to-text retrieval and text-to-image retrieval are symmetrical. And, we suppose that there is only one paired text for each image in the mini-batch. Due to the truncation operation $[x]_+$, we only discuss the case of $\mathcal{L}> 0$ that could generate gradients. Taking the image-to-text direction as an example, the gradients generated by TRL, TRL-S, and TAL are
\begin{equation}     
 \resizebox{0.86\linewidth}{!}{$                           
     \frac{\partial \mathcal{L}_{trl}}{\partial\boldsymbol{v}_i} = \hat{\boldsymbol{t}}_i-\boldsymbol{t}_i, \ \ \  \frac{\partial \mathcal{L}_{trl}}{\partial{\boldsymbol{t}}_i} = - \boldsymbol{v}_i,  \ \ \ \frac{\partial \mathcal{L}_{trl}}{\partial\hat{\boldsymbol{t}}_i} = \boldsymbol{v}_i,$}
\label{eq_g_trl} 
\end{equation}
\begin{equation}              
 \resizebox{\linewidth}{!}{$                                  
     \frac{\partial \mathcal{L}_{trls}}{\partial\boldsymbol{v}_i} = \sum^{}_{j\in\mathcal{Z}}({\boldsymbol{t}}_j-\boldsymbol{t}_i),  \frac{\partial \mathcal{L}_{trls}}{\partial{\boldsymbol{t}}_i} = - |\mathcal{Z}|\boldsymbol{v}_i,  \frac{\partial \mathcal{L}_{trls}}{\partial{\boldsymbol{t}}_j} = \boldsymbol{v}_i,$}
\label{eq_g_trls} 
\end{equation}
\begin{equation}     
 \resizebox{\linewidth}{!}{$                                           
     \frac{\partial \mathcal{L}_{tal}}{\partial\boldsymbol{v}_i} = \sum^K_{j\neq i}\beta_j (\boldsymbol{t}_j-\boldsymbol{t}_i),  \frac{\partial \mathcal{L}_{tal}}{\partial{\boldsymbol{t}}_i} = - \boldsymbol{v}_i,  \frac{\partial \mathcal{L}_{tal}}{\partial{\boldsymbol{t}}_j} = \beta_j\boldsymbol{v}_i,$}
\label{eq_g_tal} 
\end{equation}
where $\mathcal{Z}=\{z\ |\ \big[m-S(I_i,T_i)+S(I_i,T_z)\big]_+>0, z\neq i, z\in\{0,\cdots,K\} \}$, $\beta_j=\frac{\exp( \boldsymbol{v}_i^\top\boldsymbol{t}_j
/\tau)}{\sum^K_{k\neq i}\exp(\boldsymbol{v}_i^\top\boldsymbol{t}_k/\tau)} $, $\hat{\boldsymbol{t}}_i$, $\boldsymbol{t}_j$ and $\boldsymbol{t}_i$ are the hardest negative sample, negative sample, and positive sample of the anchor sample $\boldsymbol{v}_i$, respectively. Since the hardest sample is most similar to the positive one, $\frac{\partial_{trl}}{\partial\boldsymbol{v}_i}$ would easily approach $\boldsymbol{0}$ and the gradients for other negative samples except for the hardest negative one are all $\boldsymbol{0}$, which may lead to bad local minima early on in training and even cause the worst-case scenario, \ie, model collapse (see~\Cref{trl_3d}). Unlike TRL, TRL-S aims to push all negative samples away from the anchor by a constant margin and produces stronger gradients for the anchor, \ie, $\|\frac{\partial_{trls}}{\partial\boldsymbol{v}_i}\|_2\ge  \|\frac{\partial_{trl}}{\partial\boldsymbol{v}_i}\|_2$, thus avoiding model collapse (see~\Cref{trls_3d}). However, the drawback is that TRL-S treats every negative sample equally while ignoring challenging ones, which limits performance improvement. Different from TRL and TRL-S, from~\Cref{eq_g_tal}, our TAL can comprehensively consider all negative samples and exploits the anchor-negative semantic relationships to adaptively adjust the gradients for each negative, thus paying more attention to hard negatives. As a result, TAL would avoid model collapse under NC while achieving superior performance (63.35\% \textit{vs.} 44.93\% \textit{vs.} 3.64\%). More details for the derivations of gradients are provided in the supplementary material.

 \begin{algorithm}[!h]
	\renewcommand{\algorithmicrequire}{\textbf{Input:}}
	\renewcommand{\algorithmicensure}{\textbf{Output:}}
	\caption{The training process of our RDE}
	\label{Alg1}
	\begin{algorithmic}[1]
	\REQUIRE The training data $\mathcal{P}$ with $N$ image-text pairs, maximal epoch $N_e$, the cross-modal model $\mathcal{M}(\Theta)$ , and the hyper-parameters $\mathcal{R},m,\tau$;
     \STATE Initialize the backbones with the weights of the pre-trained CLIP except for the TSE module, which is randomly initialized;
	    \FOR{e = 1, 2, $\cdots$, $N_e$}
                \STATE Calculate the per-sample loss $\ell(\mathcal{M},\mathcal{P})$;
                \STATE Divide the training data with the predictions of BGE and TSE using \Cref{eq3}, respectively;
                \STATE Obtain the consensus divisions to recalibrate the correspondence labels $\{\hat{l}_{ii}\}^N_{i=1}$ with \Cref{hl_ii};
                \FOR{$\mathbf{x}$ in mini-batches $\{\mathbf{x}_m\}^M_{m=1}$ }
                    \STATE Extract the BGE and TSE features of $\mathbf{x}$;  
                    \STATE Compute the similarities between $K$ image-text pairs in $\mathbf{x}$ with above features;
                    \STATE Calculate the final matching loss $\mathcal{L}_{m}$ with \Cref{loss};                    \STATE${\Theta}=\text{Optimizer}\left(\Theta,\mathcal{L}_{m}\right)$;\\
                \ENDFOR
		\ENDFOR
		\ENSURE The optimized parameters $\hat{\Theta}$.
	\end{algorithmic}  
\label{Impl}
\end{algorithm}
\subsubsection{Training and Inference}
To train the model robustly, we use the corrected label $\hat{l}_{ii}$ instead of the original correspondence label $l_{ii}$ to compute the final matching loss, \ie,
\begin{equation}\label{loss}
    \mathcal{L}_{m} = \sum^K_{i=1} \hat{l}_{ii}(\mathcal{L}^b(I_i,T_i)+\mathcal{L}^t(I_i,T_i)),
\end{equation}
where $\mathcal{L}^b(I_i,T_i)$ and $\mathcal{L}^t(I_i,T_i)$ are the TAL losses computed by~\Cref{eqtal} with BGE and TSE similarities, respectively. The training process of RDE is shown in~\Cref{Alg1}. For the joint inference, we compute the final similarity of the image-text pair as the average of the similarities computed by both embedding modules, \ie, $S=(S^b+S^t)/2$.

\begin{table*}[t]
    \centering 
    \renewcommand{\arraystretch}{0.8}   
    \resizebox{\textwidth}{!}{
    \begin{tabular}{c|lc|ccccc|ccccc|ccccc} \toprule[1.5pt] 
        &&&\multicolumn{5}{c|}{\centering  CUHK-PEDES}&\multicolumn{5}{c|}{\centering ICFG-PEDES}&\multicolumn{5}{c}{\centering RSTPReid}\\
        Noise&\multicolumn{2}{c|}{Methods}& R-1& R-5 &R-10& mAP &mINP& R-1& R-5 &R-10& mAP &mINP& R-1& R-5 &R-10& mAP &mINP\\\midrule
         \multirow{5}{*}{0\%}&SSAN&Best&61.37&80.15&86.73&-&-&54.23& 72.63 &79.53 &- &-&43.50& 67.80& 77.15&-&-\\
         &IVT&Best&65.59&83.11&89.21&-&-&56.04& 73.60& 80.22& -& -&46.70& 70.00& 78.80 &- &-\\
         &CFine&Best&69.57&85.93&91.15&-&-&60.83& 76.55& 82.42 &- &-&50.55 &72.50 &81.60 &- &-\\
         &IRRA&Best&\underline{73.38}&\underline{89.93}&\underline{93.71}&\underline{66.13}&\underline{50.24}&\underline{63.46}&\underline{80.25}&\underline{85.82}&\underline{38.06}&\textbf{7.93}&\underline{60.20}&\underline{81.30}&\underline{88.20}&\underline{47.17}&\underline{25.28} \\
         
         &\textbf{RDE}&Best&\textbf{75.94}&\textbf{90.14}& \textbf{94.12}& \textbf{67.56}& \textbf{51.44}&\textbf{67.68} & \textbf{82.47} & \textbf{87.36} & \textbf{40.06} & \underline{7.87}&\textbf{65.35}&\textbf{83.95}&\textbf{89.90}&\textbf{50.88}&\textbf{28.08} \\

         \midrule
        \multirow{12}{*}{20\%}&\multirow{2}{*}{SSAN}&Best&46.52&68.36&77.42&42.49&28.13&40.57&62.58&71.53&20.93&2.22&35.10&60.00&71.45&28.90&12.08\\
        &&Last&45.76&67.98&76.28&40.05&24.12&40.28&62.68&71.53&20.98&2.25&33.45&58.15&69.60&26.46&10.08\\ 
        
        &\multirow{2}{*}{IVT}&Best&58.59&78.51& 85.61&57.19&45.78&50.21&69.14&76.18&34.72&\underline{8.77}&43.65&66.50&75.70&37.22&20.47\\
        &&Last&57.67&78.04&85.02&56.17&44.42& 48.70&67.42&75.06&34.44&\textbf{9.25}&37.95&63.35&73.75&34.24&19.67\\
        
        &\multirow{2}{*}{IRRA}&Best&69.74&87.09&92.20&62.28&45.84&60.76&78.26&84.01&35.87&6.80&
58.75&81.90&88.25&46.38&24.78 \\
    &&Last&69.44&87.09&92.04&62.16&45.70&60.58&78.14&84.20&35.92&6.91&54.00&77.15&85.55&43.20&22.53\\

        &\multirow{2}{*}{CLIP-C}&Best&66.41&85.15&90.89&59.36&43.02&55.25&74.76&81.32&31.09&4.94&54.45&77.80&86.70&42.58&21.38  \\
        &&Last&66.10&86.01&91.02&59.77&43.57&55.17&74.58&81.46&31.12&4.97&53.20&76.25&85.40&41.95&21.95\\

        &\multirow{2}{*}{DECL}&Best&70.29&87.04&91.93&62.84&46.54&61.95&\underline{78.36}&83.88&36.08&6.25&{61.75}&80.70&86.90&{47.70}&{26.07} \\
        &&Last&70.08&87.20&{92.14}&\underline{62.86}&46.63&61.95&\underline{78.36}&83.88&36.08&6.25&60.85&80.45&86.65&47.34&25.86 \\
        
          &\multirow{2}{*}{\textbf{RDE}}&Best& \underline{74.46}&\textbf{89.42} & \textbf{93.63} & \textbf{66.13} & \textbf{49.66} & 66.54 & \textbf{81.70} & \underline{86.70} & \underline{39.08} & {7.55}&\textbf{64.45} & \underline{83.50} & \textbf{90.00} & \underline{49.78} & \underline{27.43} \\
        &&Last&\textbf{74.53} & \underline{89.23} &\underline{93.55}& \textbf{66.13}& \underline{49.63} & \underline{66.51} & \textbf{81.70} & \textbf{86.71} & \textbf{39.09} & {7.56} &\underline{63.85}& \textbf{83.85}& \underline{89.45} & \textbf{50.27}& \textbf{27.75} \\\midrule

        \multirow{12}{*}{50\%}&\multirow{2}{*}{SSAN}&Best&13.43&31.74&41.89&14.12&6.91&18.83&37.70&47.43&9.83&1.01&19.40&39.25&50.95&15.95&6.13\\
        &&Last&11.31&28.07&37.90&10.57&3.46&17.06&37.18&47.85&6.58&0.39&14.10&33.95&46.55&11.88&4.04\\ 
        &\multirow{2}{*}{IVT}&Best&50.49&71.82&79.81&48.85&36.60&43.03&61.48&69.56&28.86&6.11&39.70&63.80&73.95&34.35&18.56 \\
        &&Last&42.02&65.04&73.72&40.49&27.89&36.57&54.83&62.91&24.30&5.08&28.55&52.05&62.70&26.82&13.97\\ 
        &\multirow{2}{*}{IRRA}&Best&62.41&82.23&88.40&55.52&38.48&52.53&71.99&79.41&29.05&4.43&56.65&78.40&86.55&42.41&21.05 \\
        &&Last&42.79&64.31&72.58&36.76&21.11&39.22&60.52&69.26&19.44&1.98&31.15&55.40&65.45&23.96&9.67 \\

        &\multirow{2}{*}{CLIP-C}&Best& 64.02&83.66&89.38&57.33&40.90&51.60&71.89&79.31&28.76&4.33&53.45&76.80&85.50&41.43&21.17\\
        &&Last& 63.97&83.74&89.54&57.35&40.88&51.49&71.99&79.32&28.77&4.37&52.35&76.35&85.25&40.64&20.45\\
        
        &\multirow{2}{*}{DECL}&Best&65.22&83.72&89.28&57.94&{41.39}&\underline{57.50}&75.09&\underline{81.24}&\underline{32.64}&\underline{5.27}&
\underline{56.75}&\underline{80.55}&\underline{87.65}&\underline{44.53}&{23.61}\\
        &&Last&65.09&83.58&89.26&57.89&41.35&57.49&\underline{75.10}&81.23&32.63&5.26&55.00&80.50&86.50&43.81&23.31\\
        &\multirow{2}{*}{\textbf{RDE}}&Best&\textbf{71.33} & \textbf{87.41} & \textbf{91.81} & \underline{63.50} & \underline{47.36} & \textbf{63.76} & \textbf{79.53} & \textbf{84.91} & \textbf{37.38} & \textbf{6.80} &\textbf{62.85} & \textbf{83.20} & \textbf{89.15} & \textbf{47.67} & \textbf{23.97}\\
        &&Last&\underline{71.25} & \underline{87.39} & \underline{91.76} & \textbf{63.59} & \textbf{47.50} & \textbf{63.76} & \textbf{79.53} & \textbf{84.91} & \textbf{37.38} & \textbf{6.80} &\textbf{62.85} & \textbf{83.20} & \textbf{89.15} & \textbf{47.67} & \underline{23.96}\\
 
        \bottomrule[1.5pt]
    \end{tabular}}
    \caption{Performance comparison under different noise rates on three benchmarks. ``Best'' means choosing the best checkpoint on the validation set to test, and ``Last'' stands for choosing the checkpoint after the last training epoch to conduct inference. R-1,5,10 is an abbreviation for Rank-1,5,10 (\%) accuracy. The best and second-best results are in \textbf{bold} and \underline{underline}, respectively.}
    \label{tabnoise}
    \vspace{-0.4cm}
\end{table*}

\section{Experiments}
In this section, we conduct extensive experiments to verify the effectiveness and superiority of the proposed RDE on three widely-used benchmark datasets.

\subsection{Datasets and Settings}
\subsubsection{Datasets} 
In the experiments, we use the CHUK-PEDES~\cite{li2017person}, ICFG-PEDES~\cite{ding2021semantically}, and RSTPReid~\cite{zhu2021dssl} datasets to evaluate our RDE. We follow the data partitions used in IRRA~\cite{jiang2023cross} to split the datasets into training, validation, and test sets, wherein the ICFG-PEDES dataset only has training and validation sets. More details are provided in the supplementary material.

\subsubsection{Evaluation Protocols}
For all experiments, we mainly employ the popular Rank-K metrics (K=1,5,10) to measure the retrieval performance. In addition to Rank-K, we also adopt the mean Average Precision (mAP) and mean Inverse Negative Penalty (mINP) as auxiliary retrieval metrics to further evaluate performance following \cite{jiang2023cross}.

\subsubsection{Implementation Details}
As mentioned earlier, we adopt the pre-trained model CLIP~\cite{radford2021learning} as our modality-specific encoders. In fairness, we use the same version of CLIP-ViTB/16 as IRRA~\cite{jiang2023cross} to conduct experiments. During training, we introduce data augmentations to increase the diversity of the training data. Specifically, we utilize random horizontal flipping, random crop with padding, and random erasing to augment the training images. For training texts, we employ random masking, replacement, and removal for the word tokens as the data augmentation.
Moreover, the input size of images is $384\times128$ and the maximum length of input word tokens is set to 77. We employ the Adam optimizer to train our model for 60 epochs with a cosine learning rate decay strategy. The initial learning rate is $1e-5$ for the original model parameters of CLIP and the initial one for the network parameters of TSE is initialized to $1e-3$. The batch size is 64. Following IRRA~\cite{jiang2023cross}, we adopt an early training process with a gradually increasing learning rate. For hyperparameter settings, the margin value $m$ of TAL is set to 0.1, the temperature parameter $\tau$ is set to 0.015, and the selection ratio $\mathcal{R}$ is 0.3. 

\subsection{Comparison with State-of-the-Art Methods}
In this section, we evaluate the performance of our RDE on three benchmarks under different scenarios. For a comprehensive comparison, we compare our method with several state-of-the-art methods, including both ordinary methods and robust methods. Moreover, we use two synthetic noise levels (\ie, noise rates), 20\%, and 50\%, to simulate the real-world scenario where the image-text pairs are not well-aligned. We randomly shuffle the text descriptions to inject NCs into the training data. We compare our RDE with five state-of-the-art baselines:  SSAN~\cite{ding2021semantically}, IVT~\cite{shu2022see}, IRRA~\cite{jiang2023cross}, DECL~\cite{qin2022deep}, and CLIP-C. SSAN, IVT, and IRRA are recent ordinary methods that are not designed for NC. DECL is a general framework that can enhance the robustness of image-text matching methods against NC. We use the model of IRRA as the base model of DECL for TIReID. CLIP-C is a strong baseline that fine-tunes the CLIP(ViT-B/16) model with only clean image-text pairs. We report the results of both the best checkpoint on the validation set and the last checkpoint to show the overfitting degree. Furthermore, we also evaluate our RDE on the original datasets without synthetic NC to demonstrate its superiority in \Cref{tabnoise}. We compare our RDE with two local-matching methods: SSAN~\cite{ding2021semantically} and CFine~\cite{yan2022clip}); and two global-matching methods: IVT~\cite{shu2022see} and IRRA~\cite{jiang2023cross}. More comparisons with other methods are provided in the supplementary material.

From~\Cref{tabnoise}, one can see that our RDE  achieves state-of-the-art performance on three datasets and we can draw three observations: \textcolor{red}{{(1)}} On the datasets with synthetic NC, the ordinary methods suffer from remarkable performance degradation or poor performance as the noise rate increases. In contrast, our RDE achieves the best results on all metrics. Moreover, by comparing the `Best' performance with the `Last' ones in~\Cref{tabnoise}, we can see that our RDE can effectively prevent the performance deterioration caused by overfitting against NC. \textcolor{red}{{(2)}} Compared with the robust framework DECL and the strong baseline CLIP-C, our RDE also shows obvious advantages, which indicates that our solution against NC is effective and superior in TIReID. For instance, on CUHK-PEDES under 50\% noise, our RDE achieves 71.33\%, 87.41\%, and 91.81\% in terms of Rank-1,5,10 on the `Best' rows, respectively, which surpasses the best baseline DECL by a large margin, \ie, +6.11\%, +3.69\%, and +2.53\%, respectively. \textcolor{red}{{(3)}} On the datasets without synthetic NC, our RDE outperforms all baselines by a large margin. Specifically, RDE achieves performance gains of +2.56\%, +4.22\%, and +5.15\% in terms of Rank-1 compared with the best baseline IRRA on three datasets, respectively, demonstrating the effectiveness and advantages of our method.

\subsection{Ablation Study}
In this section, we conduct ablation studies on the CUHK-PEDES dataset with 50\% noise to investigate the effects and contributions of each proposed component in RDE. We compare different combinations of our components in \Cref{tab_ab1}. From the experimental results, we could draw the following observation: \textcolor{red}{{(1)}} RDE achieves the best performance by using both BGE and TSE for joint inference, which demonstrates that these two modules are complementary and effective. \textcolor{red}{{(2)}} RDE benefits from CCD, which can enhance the robustness and alleviate the overfitting effect caused by NC. \textcolor{red}{{(3)}} Our TAL outperforms the widely-used Triplet Ranking Loss (TRL) and SDM loss~\cite{jiang2023cross}, which demonstrates the superior stability and robustness of our TAL against NC. 
\begin{table}[h]
\centering
\vspace{-0.2cm}
\renewcommand{\arraystretch}{0.9}   
\setlength\tabcolsep{4.pt}
\resizebox{\linewidth}{!}{
\begin{tabular}{ccccc|ccccc}\toprule[1.2pt]
No.&$S^b$&$S^e$&CCD&Loss&R-1&R-5&R-10&mAP&mINP \\\midrule
\#1&\checkmark&\checkmark&\checkmark&TAL&\textbf{71.33}& \textbf{87.41}& \textbf{91.81}& \textbf{63.50}& \textbf{47.36} \\
\#2&\checkmark&\checkmark&\checkmark&TRL&6.40 & 16.08 & 22.14 & 6.53 & 2.51\\
\#3&\checkmark&\checkmark&\checkmark&TRL-S& 67.38& 85.35& 90.64& 60.04& 43.60   \\
\#4&\checkmark&\checkmark&\checkmark&SDM&69.33 & 86.99 & 91.68 & 61.99 & 45.34\\
\#5&&\checkmark&\checkmark&TAL&70.70 & 86.60 & 91.16 & 62.67 & 46.19\\
\#6&\checkmark&&\checkmark&TAL&69.07 & 86.09 & 91.13 & 61.69 & 45.40\\
\#7&\checkmark&\checkmark&&TAL& 63.11&81.04&87.22&55.42&38.68\\
\bottomrule[1.2pt]
    \end{tabular}}
    \caption{Ablation studies on the CHUK-PEDES dataset.}
    \label{tab_ab1}
\vspace{-0.5cm}
\end{table}

\subsection{Parametric Analysis}
\begin{figure}[h]
\centering
\begin{subfigure}{0.49\linewidth}
\includegraphics[width=1\linewidth]{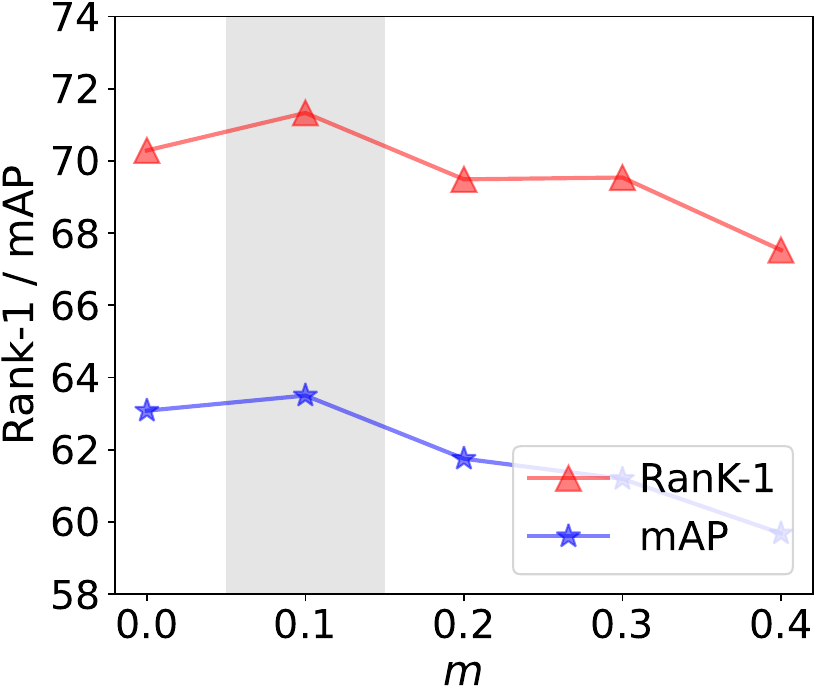}
\centering
\end{subfigure}
\begin{subfigure}{0.49\linewidth}
\hfill
\includegraphics[width=1\linewidth]{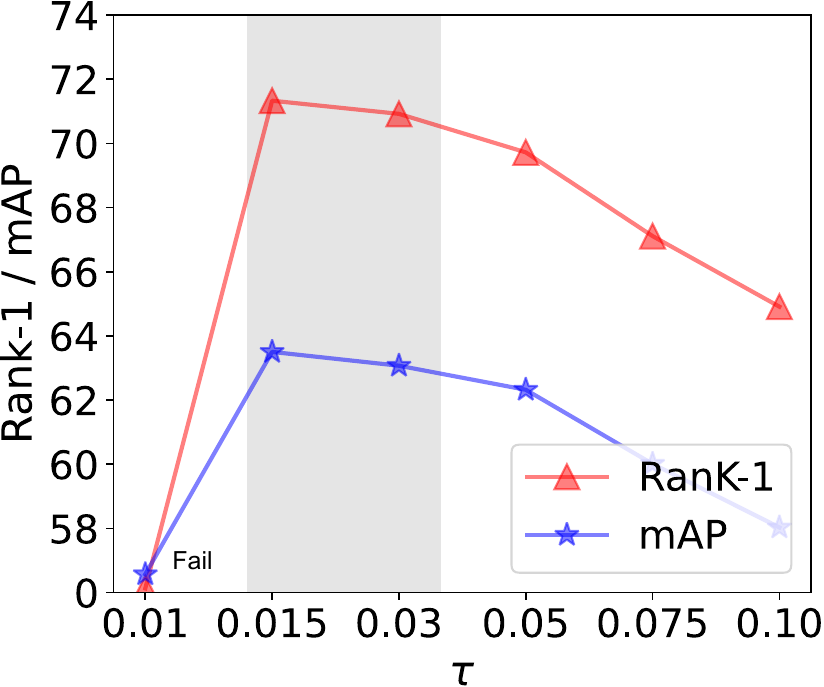}
\centering
\end{subfigure}
\setlength{\abovecaptionskip}{0.1cm}
\caption{Variation of performance with different $m$ and $\tau$.}
\label{figp}
\vspace{-0.4cm}
\end{figure}

To study the impact of different hyperparameter settings on performance,  we perform sensitivity analyses for two key hyperparameters  (\ie, $m$ and $\tau$) on the CHUK-PEDES dataset with 50\% noise.  From~\Cref{figp}, we can see that: \textcolor{red}{{(1)}} Too large or too small $m$ will lead to suboptimal performance. We choose $m=0.1$ in all our experiments. \textcolor{red}{{(2)}} Too small $\tau$ will cause training failure, while the increasing $\tau$ will gradually decrease the separability (hardness) of positive and negative pairs for suboptimal performance. We choose $\tau=0.015$ in all our experiments.


\subsection{Robustness Study}
\begin{figure}[h]
\centering
\begin{subfigure}{0.49\linewidth}
\includegraphics[width=1\linewidth]{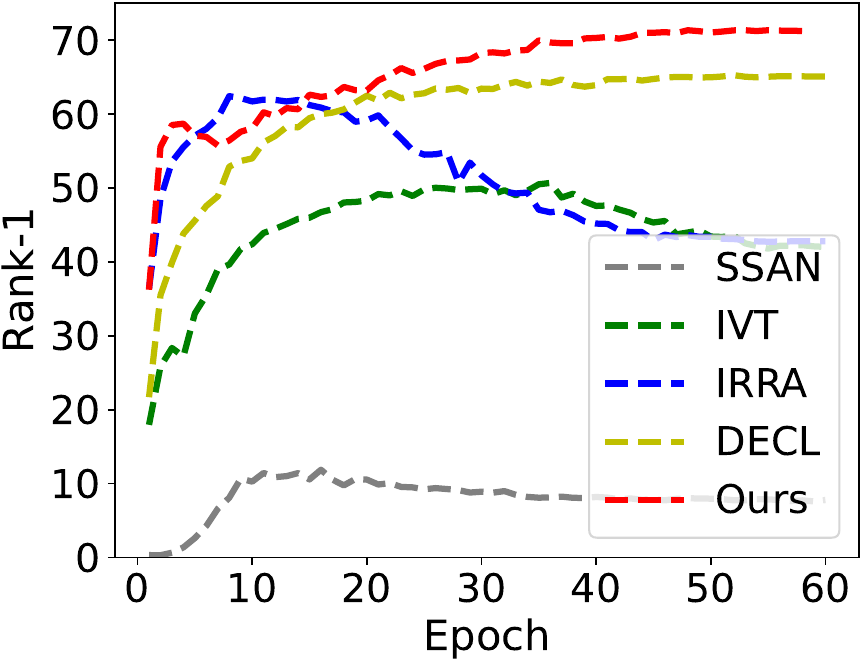}
\centering
\caption{CUHK-PEDES} 
\end{subfigure}
\begin{subfigure}{0.49\linewidth}
\hfill
\includegraphics[width=1\linewidth]{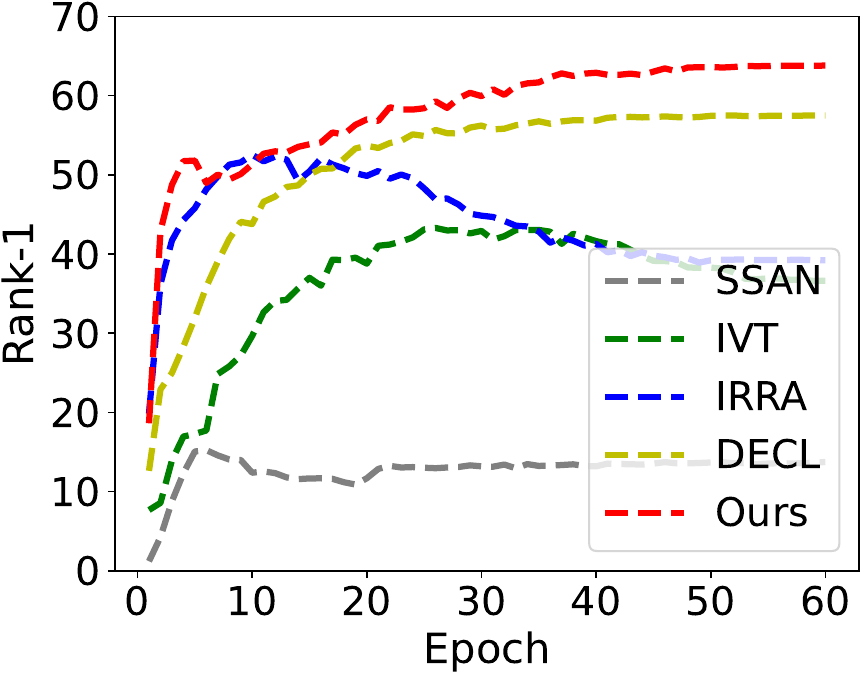}
\centering
\caption{ICFG-PEDES}
\end{subfigure}
\setlength{\abovecaptionskip}{0.1cm}
\caption{Test performance (Rank-1) versus epochs on the CHUK-PEDES and ICFG-PEDES datasets with 50\% noise. }
\label{fig4} 
\vspace{-0.4cm}
\end{figure}

In this section, we provide some visualization results during cross-modal training to verify the robustness and effectiveness of our method. As shown in \Cref{fig4}, one can clearly see that our RDE not only achieves excellent performance under noise but also effectively alleviates noise overfitting.

\section{Conclusion}
In this paper, we reveal and study a novel challenging problem of noisy correspondence (NC) in TIReID, which violates the common assumption of existing methods that image-text data is perfectly aligned. To this end, we propose a robust method, \ie, RDE, to effectively handle the revealed NC problem and achieve superior performance. Extensive experiments are conducted on three datasets to comprehensively demonstrate the superiority and robustness of RDE both with and without synthetic NCs.

\section*{Acknowledgments}

This work was supported in part by NSFC under Grant U21B2040, 62176171, 62372315, and 62102274, in part by Sichuan Science and Technology Program under Grant 2022YFH0021 and 2023ZYD0143; in part by Chengdu Science and Technology Project under Grant 2023-XT00-00004-GX; in part by the SCU-LuZhou Sciences and Technology Coorperation Program under Grant 2023CDLZ-16; in part by the Fundamental Research Funds for the Central Universities under Grant CJ202303 and YJ202140.


\begin{figure*}[t]
    \centering
    \resizebox{\linewidth}{!}{ 
    \includegraphics{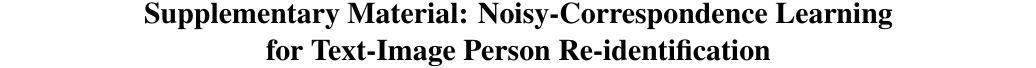}}
    \vspace{1.5cm}
\end{figure*}
\appendix



In this supplementary material, we provide additional information for RDE. More specifically, we first give detailed proof and derivation for lemmas and gradients in~\Cref{app1}. In~\Cref{app2}, we detail the used datasets and the compared baselines. In~\Cref{app3}, to further verify the robustness of RDE, we provide the re-identification performance on three benchmark datasets under extremely high noise rate, \ie, 80\%. Besides, in~\Cref{app4}, we provide more comparison results compared with state-of-the-art methods to comprehensively verify the superiority of our RDE. In~\Cref{app5}, we explore the impact of different selection ratios ($\mathcal{R}$) on performance. In~\Cref{app6}, we provide a more ablation analysis. In~\Cref{app7}, we provide a large number of real noisy examples existing in the three public datasets to conduct a case study, thus emphasizing our motivation. We also provide a more comprehensive robustness analysis to verify the robustness of RDE in~\Cref{app8}. Finally, in~\Cref{app9}, we provide some qualitative results to illustrate the advantages of our RDE.

\section{Proof and Derivation\label{app1}}
\subsection{Proof for~\Cref{lam1}(\Cref{lam1_})}
Given an input image-text pair $(I_i,T_i)$ in a mini-batch $\mathbf{x}$, TAL is defined as:
\begin{equation}
 \resizebox{\linewidth}{!}{$
    \begin{aligned}
        \mathcal{L}_{tal}(I_i,T_i)&=\big[ m -  S^+_{i2t}(I_i) + \tau \log ( \sum_{j = 1}^K q_{ij}\exp(S(I_i,T_j)/\tau)  ) \big]_+\\ &+\big[ m -  S^+_{t2i}(T_i)  + \tau \log ( \sum_{j = 1}^K q_{ji}\exp(S(I_j,T_i)/\tau)  ) \big]_+,
    \end{aligned}$}
    \label{eqtal}
\end{equation}
where $m$ is a positive margin coefficient, $\tau$ is a temperature coefficient to control hardness, $S(I_i,T_j)\in \{S_{ij}^b, S_{ij}^t\}$, $[x]_+ \equiv \max (x,0)$, $\exp(x)\equiv e^x$, $q_{ij}=1-l_{ij}$,  and $K$ is the size of $\mathbf{x}$. From~\Cref{lam1}, as $\tau\to0$, TAL is close to TRL and focuses more on hard negatives.  Since multiple positive pairs from the same identity may appear in the mini-batch, $S^+_{i2t}(I_i) =\sum^K_{j=1} \alpha_{ij}  S(I_i,T_j)$ is the weighted average similarity of positive pairs for image $I_i$, where $\alpha_{ij}=\frac{l_{ij}\exp{(S(I_i,T_j)/\tau)}}{\sum^N_{k=1}l_{ik}\exp{(S(I_i,T_k)/\tau)}}$. And, $S^+_{i2t}(T_i)$ is similar to the definition of $S^+_{i2t}(I_i)$.  
\begin{lemma}
    TAL is the upper bound of TRL, i.e.,
    \begin{equation}
    \begin{aligned}
        &\mathcal{L}_{trl}(I_i,T_i)= \big[ m -  S^+_{i2t}(I_i) +  S(I_i,\hat{T}_i)  \big]_+    \\
&+\big[ m -  S^+_{t2i}(T_i)  + S(\hat{I}_i,T_i) \big]_+ \leq \mathcal{L}_{tal}(I_i,T_i),
    \end{aligned}
    \label{eq2}
    \end{equation}
    where $\hat{T}_i\in\mathbf{T}_i=\{ T_j|l_{ij}=0, \forall j\in \{1,2,\cdots,K\} \}$ is the hardest negative text for image $I_i$ and $\hat{I}_i\in\mathbf{I}_i=\{ I_j|l_{ji}=0, \forall j\in \{1,2,\cdots,K\} \}$ is the hardest negative image for text $I_i$, respectively. 
    \label{lam1_}

    \begin{proof}
To prove~\Cref{eq2}, we first take the image-to-text direction as an example. For $S(I_i,\hat{T}_i)$ in~\Cref{eq2}, we have that
\begin{equation}
    \begin{aligned}
            S(I_i,\hat{T}_i) &= \max_{T_j\in\mathbf{T}_i}\left(S(I_i,T_j)\right)\\
            &=\max_{T_j\in\mathbf{T}_i}\left( \tau \log \exp\left(S(I_i,T_j)\right)^{\frac{1}{\tau}} \right)\\
             &= \tau \log \left(\max_{T_j\in\mathbf{T}_i} \left( \exp\left(S(I_i,T_j)\right)^{\frac{1}{\tau}} \right) \right)\\
             &\leq \tau \log \left(  \sum_{T_j\in\mathbf{T}_i} \exp(S(I_i,T_j)/\tau)\right)\\
             &\leq \tau \log (  \sum_{j=1}^K q_{ij} \exp(S(I_i,T_j)/\tau)),
    \end{aligned}
    \label{eq3}
    \end{equation}
    where $q_{ij} = 1 - l_{ij}$. Based on~\Cref{eq3}, we have that
    \begin{equation}
    \begin{gathered}
        \big[ m -  S^+_{i2t}(I_i) + \tau \log ( \sum_{j = 1}^K q_{ij}\exp(S(I_i,T_j)/\tau)  ) \big]_+\\ \ge
        \big[ m -  S^+_{i2t}(I_i) +  S(I_i,\hat{T}_i)  \big]_+
        .
    \end{gathered}
    \label{eq4}
    \end{equation}
    Similarly, in the text-to-image direction, we have that
    \begin{equation}
    \begin{gathered}
        \big[ m -  S^+_{t2i}(T_i)  + \tau \log ( \sum_{j = 1}^K q_{ji}\exp(S(I_j,T_i)/\tau)  ) \big]_+\\
        \ge \big[ m -  S^+_{t2i}(T_i)  + S(\hat{I}_i,T_i) \big]_+.
    \end{gathered}
    \label{eq5}
    \end{equation}
    Thus, combining~\Cref{eq4} and~\Cref{eq5}, we can get $\mathcal{L}_{trl}(I_i,T_i)\leq\mathcal{L}_{tal}(I_i,T_i)$.  This completes the proof.
    \end{proof} 
    
\end{lemma}

\subsection{Derivation for Gradient}
In this appendix, we provide more details of gradient derivation. For ease of representation and analysis, we only consider one direction like~\cite{li2023selectively} since image-to-text retrieval and text-to-image retrieval are symmetrical. Besides, we suppose that there is only one paired text for each image in the mini-batch. Thus, TRL, TRL-S, and TAL are simplified as follows:
 \begin{equation}
     \begin{gathered}
        \mathcal{L}_{trl}(I_i,T_i)= \big[ m- \boldsymbol{v}_i^\top\boldsymbol{t}_i  + \boldsymbol{v}_i^\top\hat{\boldsymbol{t}}_i \big]_+,\\
         \mathcal{L}_{trls}(I_i,T_i)= \sum^K_{j\neq i} \big[ m- \boldsymbol{v}_i^\top\boldsymbol{t}_i  + \boldsymbol{v}_i^\top{\boldsymbol{t}}_j  \big]_+,\\
         \mathcal{L}_{tal}(I_i,T_i)= \Big[ m- \boldsymbol{v}_i^\top\hat{\boldsymbol{t}}_i +\tau\log(\sum^K_{j\neq i}e^{(\boldsymbol{v}_i^\top{\boldsymbol{t}}_j/\tau)})\Big]_+,
     \end{gathered}
 \end{equation}
where $\hat{\boldsymbol{t}}_i$, $\boldsymbol{t}_j$ and $\boldsymbol{t}_i$ are the hardest negative sample, negative sample, and positive sample of the anchor sample $\boldsymbol{v}_i$, respectively. These $\ell_2$-normalized features are embedded by the modality-specifical models, \ie, $f_{\Theta_{v}}(\cdot)$ and $f_{\Theta_{t}}(\cdot)$. Due to the truncation operation $[x]_+$, we only discuss the case of $\mathcal{L} > 0$ that could generate gradients. For TRL, the gradients to the parameters $\Theta_v$ and $\Theta_t$ are:
\begin{equation}
    \begin{gathered}
        \frac{\partial\mathcal{L}_{trl}}{\partial\Theta_v}=\frac{\partial\mathcal{L}_{trl}}{\partial\boldsymbol{v}_i}\frac{\partial\boldsymbol{v}_i}{\partial\Theta_v},\\
        \frac{\partial\mathcal{L}_{trl}}{\partial\Theta_t}=\frac{\partial\mathcal{L}_{trl}}{\partial\hat{\boldsymbol{t}}_i}\frac{\partial\hat{\boldsymbol{t}}_i}{\partial\Theta_t}+\frac{\partial\mathcal{L}_{trl}}{\partial{\boldsymbol{t}}_i}\frac{\partial{\boldsymbol{t}}_i}{\partial\Theta_t}.
    \end{gathered}
\end{equation}
Since the learning of normalized features can be viewed as the movement process of points on a unit hyperplane, we only consider the loss gradients with respect to $\boldsymbol{v}_i$, $\hat{\boldsymbol{v}}_i$, and $\boldsymbol{t }_i$ are:
\begin{equation}                              
     \frac{\partial \mathcal{L}_{trl}}{\partial\boldsymbol{v}_i} = \hat{\boldsymbol{t}}_i-\boldsymbol{t}_i,  \frac{\partial \mathcal{L}_{trl}}{\partial{\boldsymbol{t}}_i} = - \boldsymbol{v}_i,  \frac{\partial \mathcal{L}_{trl}}{\partial\hat{\boldsymbol{t}}_i} = \boldsymbol{v}_i.
\label{eq_g_trl} 
\end{equation}
For TRL-S, the gradients to the parameters $\Theta_v$ and $\Theta_t$ are:
\begin{equation}
    \begin{gathered}
        \frac{\partial\mathcal{L}_{trls}}{\partial\Theta_v}=\frac{\partial\mathcal{L}_{trls}}{\partial\boldsymbol{v}_i}\frac{\partial\boldsymbol{v}_i}{\partial\Theta_v},\\
        \frac{\partial\mathcal{L}_{trls}}{\partial\Theta_t}=\sum_{j\in \mathcal{Z}}\frac{\partial\mathcal{L}_{trls}}{\partial{\boldsymbol{t}}_j}\frac{\partial{\boldsymbol{t}}_j}{\partial\Theta_t}+\frac{\partial\mathcal{L}_{trls}}{\partial{\boldsymbol{t}}_i}\frac{\partial{\boldsymbol{t}}_i}{\partial\Theta_t}.
    \end{gathered}
\end{equation}
Thus, for $\boldsymbol{v}_i$, ${\boldsymbol{v}}_j$, and $\boldsymbol{t }_i$, the gradients are:
\begin{equation}              
  \begin{gathered}
       \frac{\partial \mathcal{L}_{trls}}{\partial\boldsymbol{v}_i} = \sum^{}_{j\in\mathcal{Z}}({\boldsymbol{t}}_j-\boldsymbol{t}_i),   \frac{\partial \mathcal{L}_{trls}}{\partial{\boldsymbol{t}}_j} = \boldsymbol{v}_i, \forall j\in \mathcal{Z},\\ \frac{\partial \mathcal{L}_{trls}}{\partial{\boldsymbol{t}}_i} = - \sum_{j\in\mathcal{Z}}\boldsymbol{v}_i = -|\mathcal{Z}|\boldsymbol{v}_i, 
  \end{gathered}
\label{eq_g_trls} 
\end{equation}
where $\mathcal{Z}=\{z\ |\ \big[m-S(I_i,T_i)+S(I_i,T_z)\big]_+>0, z\neq i, z\in\{0,\cdots,K\} \}$. For our TAL, the gradients to the parameters $\Theta_v$ and $\Theta_t$ are:
\begin{equation}
    \begin{gathered}
        \frac{\partial\mathcal{L}_{tal}}{\partial\Theta_v}=\frac{\partial\mathcal{L}_{tal}}{\partial\boldsymbol{v}_i}\frac{\partial\boldsymbol{v}_i}{\partial\Theta_v},\\
        \frac{\partial\mathcal{L}_{tal}}{\partial\Theta_t}=\sum_{j\neq i}\frac{\partial\mathcal{L}_{tal}}{\partial{\boldsymbol{t}}_j}\frac{\partial{\boldsymbol{t}}_j}{\partial\Theta_t}+\frac{\partial\mathcal{L}_{tal}}{\partial{\boldsymbol{t}}_i}\frac{\partial{\boldsymbol{t}}_i}{\partial\Theta_t}.
    \end{gathered}
\end{equation}
Thus,  the gradients  for $\boldsymbol{v}_i$, ${\boldsymbol{v}}_j$ $\boldsymbol{t }_i$ are:
\begin{equation}                          
\begin{gathered}
     \frac{\partial \mathcal{L}_{tal}}{\partial\boldsymbol{v}_i} = \sum^K_{j\neq i}\beta_j\boldsymbol{t}_j -\boldsymbol{t}_i  = \sum^K_{j\neq i}\beta_j (\boldsymbol{t}_j-\boldsymbol{t}_i), 
     \\
     \frac{\partial \mathcal{L}_{tal}}{\partial{\boldsymbol{t}}_i} = - \boldsymbol{v}_i,\ \ \   \frac{\partial \mathcal{L}_{tal}}{\partial{\boldsymbol{t}}_j} = \beta_j\boldsymbol{v}_i, 
\end{gathered}
\label{eq_g_tal} 
\end{equation}
where $\beta_j=\frac{\exp( \boldsymbol{v}_i^\top\boldsymbol{t}_j
/\tau)}{\sum^K_{k\neq i}\exp(\boldsymbol{v}_i^\top\boldsymbol{t}_k/\tau)}$.
\begin{table*}[t]
\setlength\tabcolsep{5.pt}
    \centering
    \resizebox{\textwidth}{!}{
    \begin{tabular}{c|lc|ccccc|ccccc|ccccc} \toprule[1pt] 
        &&&\multicolumn{5}{c|}{\centering  CUHK-PEDES}&\multicolumn{5}{c|}{\centering ICFG-PEDES}&\multicolumn{5}{c}{\centering RSTPReid}\\
        Noise&\multicolumn{2}{c|}{Methods}& R-1& R-5 &R-10& mAP &mINP& R-1& R-5 &R-10& mAP &mINP& R-1& R-5 &R-10& mAP &mINP\\\midrule
        \multirow{12}{*}{80\%}&\multirow{2}{*}{SSAN}&Best&0.18&0.83&1.45&0.47&0.24&0.28&0.99&1.90&0.27&0.15&0.65&3.25&5.95&1.30&0.70

        \\
        &&Last&0.13&0.67&1.46&0.42&0.21&0.18&1.01&1.77&0.25&0.14&0.65&2.95&5.85&1.32&0.68\\ 
        &\multirow{2}{*}{IVT}&Best&34.03&55.49&66.16&33.90&23.29&21.10&37.10&45.64&13.68&2.32&15.15&30.00&40.50&14.98&7.79\\
        &&Last&10.61&23.81&31.38&11.13&5.72&5.64&12.48&17.15&4.00&0.69&4.95&13.55&19.75&6.07&2.85\\ 
        &\multirow{2}{*}{IRRA}&Best&38.63&56.69&64.18&34.60&21.84&28.19&44.14&51.27&14.36&1.41&29.65&46.65&54.50&23.77&11.32 \\
        &&Last&9.06&19.69&25.65&8.26&3.18&8.68&18.76&24.50&3.65&0.27&8.15&21.00&29.05&7.28&2.40\\

        &\multirow{2}{*}{CLIP-C}&Best&57.38&78.05&84.97&51.08&{34.83}&44.84&65.24&73.27&{24.27}&3.42& 47.80&{72.70}&{81.75}&{37.50}&18.09\\
        &&Last&57.05&78.09&85.07&51.14&\underline{35.05}&44.65&65.26&{73.45}&24.20&{3.44}&44.60&70.75&80.20&35.67&17.09\\
        
        &\multirow{2}{*}{DECL}&Best&47.90&71.57&80.17&44.51&29.86 &40.53&61.49&69.84&21.78&2.97&{48.15}&72.20&80.75&37.31&\textbf{18.83} \\
        &&Last&46.57&70.19&78.48&42.93&27.91&39.91&61.16&69.51&21.56&2.89&45.85&71.05&81.00&35.34&16.35\\
        &\multirow{2}{*}{\textbf{RDE}}&Best&\textbf{64.99} & \underline{83.15} & \underline{89.52} & \textbf{57.84} & \textbf{41.07}&\textbf{56.02} & \underline{74.00} &\textbf{80.62} & \underline{30.67} & \underline{4.60}&\textbf{53.40} & \underline{76.70} & \underline{85.55} & \underline{39.71} & \underline{18.28} \\
        
        &&Last&\underline{64.91} & \textbf{83.20} & \textbf{89.54} & \underline{57.83} & \textbf{41.07}&
\underline{55.96} & \textbf{74.09} & \underline{80.61} & \textbf{30.79} & \textbf{4.62}&\underline{52.35} & \textbf{76.85} & \textbf{84.90} & \textbf{39.92} & {17.72}\\

        \bottomrule[1pt]
    \end{tabular}}
\setlength\tabcolsep{6.pt}
    \caption{Performance comparison under 80\% noise rate on three benchmarks. ``Best'' means choosing the best checkpoint on the validation set to test, and ``Last'' stands for choosing the checkpoint after the last training epoch to conduct inference. R-1,5,10 is an abbreviation for Rank-1,5,10 (\%) accuracy. The best and second-best results are in \textbf{bold} and \underline{underline}, respectively.}
    \label{tab80noise}
\end{table*}
\section{Dataset and Baseline Description \label{app2}}
\subsection{Datasets.} To verify the effectiveness and superiority of RDE, we use three widely-used image-text person datasets to conduct experiments. A brief introduction of these datasets is given as follows:
\begin{itemize}
        \item \textbf{CHUK-PEDES}~\cite{li2017person} is the first large-scale benchmark to dedicate TIReID, which includes 40,206 person images and 80,412 text descriptions for 13,003 unique identities. We follow the official data split to conduct experiments, \ie, 11,003 identities for training, 1,000 identities for validation, and the rest of the 1,000 identities for testing.
        \item \textbf{ICFG-PEDES}~\cite{ding2021semantically} is a widely-used benchmark collected from the MSMT17 dataset~\cite{wei2018person} and consists of 54,522 images for 4,102 unique persons and each image has a corresponding textual description. We follow the data split used by most TIReID methods~\cite{shu2022see,jiang2023cross}, \ie, a training set with 3,102 identifies and a validation set with 1,000 identifies. Note that we uniformly used the validation performance as the test performance due to its lack of a test set.
        \item  \textbf{RSTPReid}~\cite{zhu2021dssl} is another benchmark dataset constructed from the MSMT17 dataset~\cite{wei2018person} for TIReID. RSTPReid contains 20,505 images for 4,101 identities, wherein each person has 5 images and each image is paired with 2 text descriptions. Following the official data split, we use 3,701 identities for training, 200 identities for validation, and the remaining 200 identities for testing.
\end{itemize}

\subsection{Baselines.} To verify the effectiveness and robustness of our method in the NC scenario, we provide the comparison results with 5 baselines that have published code. We introduce each baseline as follows:
\begin{itemize}
    \item \textbf{SSAN}\footnote{https://github.com/zifyloo/SSAN}~\cite{ding2021semantically} is a local-matching method for TIReID, which mainly benefits from a proposed multiview non-local network that could capture the local relationships, thus establishing better correspondences between body parts and noun phrases. Besides, SSAN also exploits a compound ranking loss to make an effective reduction of the intra-class variance in textual features.
    \item \textbf{IVT}\footnote{https://github.com/TencentYoutuResearch/PersonRetrieval-IVT}~\cite{shu2022see} is an implicit visual-textual framework, which belongs to the global-matching method. To explore fine-grained alignments, IVT utilizes two implicit semantic alignment paradigms, \ie, multi-level alignment (MLA) and bidirectional mask modeling (BMM). MLA aims to see “finer” by exploring local and global alignments from three-level matchings. BMM aims to see “more” by mining more semantic alignments from random masking for both modalities.
    \item \textbf{IRRA}\footnote{https://github.com/anosorae/IRRA}~\cite{jiang2023cross} is a recent state-of-art global-matching method that could learn relations between local visual-textual tokens and enhances global alignments without requiring additional prior supervision. IRRA exploits a novel similarity distribution matching to minimize the KL divergence between the similarity distributions and the normalized label matching distributions for better performance.
    \item \textbf{CLIP-C} is a quite strong baseline that fine-tunes the original CLIP\footnote{https://github.com/openai/CLIP} model with only clean image-text pairs. We use the same version as IRRA, \ie, ViTB/16, for a fair comparison and use InfoNCE loss~\cite{oord2018representation} to optimize the model.
    \item \textbf{DECL}\footnote{https://github.com/QinYang79/DECL}~\cite{qin2022deep} is an effective robust image-text matching framework, which utilizes the cross-modal evidential learning paradigm to capture and leverage the uncertainty brought by noise to isolate the noisy pairs. Since TIReID can be treated as the sub-task of instance-level image-text matching, DECL also can be used to ease the negative impact of NCs in TIReID. In this paper, we exploit the used model of IRRA~\cite{jiang2023cross} as the base model of DECL for robust TIReID. 
\end{itemize}

\section{The Results under Extreme Noise\label{app3}}
To further verify the effectiveness and robustness of our method, we report comparison results under extremely high noise, \ie, 80\%. From the results in~\Cref{tab80noise}, one can see that our RDE achieves the best performance and can effectively alleviate the performance degradation caused by noise overfitting. For example, compared with the ‘Best’ rows, our RDE surpasses the best baselines by +7.56\%, +5.95\%, and +3.5\%  in terms of Rank-1 on the three datasets, respectively.

\section{More Comparisons \label{app4}}
In this section, we follow the organization of IRRA~\cite{jiang2023cross} and provide more comparative experimental results on three benchmarks in~\Cref{tabcuhk,tabicfg,tabrstp}. From the results, our RDE achieves the best results and exceeds the best baselines, \ie, +0.92\%, +2.63\%, and +0.15\% in terms of Rank-1 on three datasets, respectively.

\begin{table*}[t]
\centering
\resizebox{0.75\textwidth}{!}{
\begin{tabular}{l|lcc|ccccc}\toprule[1pt] 
Methods            & Ref.      & Image Enc. & Text Enc.    & R-1 & R-5 & R-10 & mAP   & mINP  \\\midrule
CMPM/C~\cite{zhang2018deep}& ECCV'18   & RN50       & LSTM         & 49.37  & -      & 79.27   & -     & -     \\
TIMAM~\cite{sarafianos2019adversarial}              & ICCV'19   & RN101      & BERT         & 54.51  & 77.56  & 79.27   & -     & -     \\
ViTAA~\cite{wang2020vitaa}              & ECCV'20   & RN50       & LSTM         & 54.92  & 75.18  & 82.90   & 51.60 & -     \\
NAFS~\cite{gao2021contextual}               & arXiv'21  & RN50       & BERT         & 59.36  & 79.13  & 86.00   & 54.07 & -     \\
DSSL~\cite{zhu2021dssl}               & MM'21     & RN50       & BERT         & 59.98  & 80.41  & 87.56   & -     & -     \\
SSAN~\cite{ding2021semantically}               & arXiv'21  & RN50       & LSTM         & 61.37  & 80.15  & 86.73   &       & -     \\
Lapscore~\cite{wu2021lapscore}           & ICCV'21   & RN50       & BERT         & 63.40   & -      & 87.80   & -     & -     \\
ISANet~\cite{yan2022image}             & arXiv'22  & RN50       & LSTM         & 63.92  & 82.15  & 87.69   & -     & -     \\
LBUL~\cite{wang2022look}               & MM'22     & RN50       & BERT         & 64.04  & 82.66  & 87.22   & -     & -     \\
\citeauthor{han2021text}2021 & BMVC'21   & CLIP-RN101 & CLIP-Xformer & 64.08  & 81.73  & 88.19   & 60.08 & -     \\
SAF~\cite{li2022learning}                & ICASSP'22 & ViT-Base   & BERT         & 64.13  & 82.62  & 88.40   & -     & -     \\
TIPCB~\cite{chen2022tipcb}              & Neuro'22  & RN50       & BERT         & 64.26  & 83.19  & 89.10   & -     & -     \\
CAIBC~\cite{wang2022caibc}              & MM'22     & RN50       & BERT         & 64.43  & 82.87  & 88.37   & -     & -     \\
AXM-Net~\cite{farooq2022axm}            & MM'22     & RN50       & BERT         & 64.44  & 80.52  & 86.77   & 58.73 & -     \\
LGUR~\cite{shao2022learning}               & MM'22     & DeiT-Small & BERT         & 65.25  & 83.12  & 89.00   & -     & -     \\
IVT~\cite{shu2022see}                & ECCVW'22  & ViT-Base   & BERT         & 65.59  & 83.11  & 89.21   & -     & -     \\
CFine~\cite{yan2022clip}              & TIP'23  & CLIP-ViT   & BERT         & 69.57  & 85.93  & 91.15   & -     & -     \\
IRRA~\cite{jiang2023cross}               & CVPR'23   & CLIP-ViT   & CLIP-Xformer & 73.38  & 89.93  & 93.71   & 66.13 & 50.24 \\
BiLMa~\cite{fujii2023bilma}    & ICCVW'23   & CLIP-ViT   & CLIP-Xformer  &74.03& 89.59& 93.62 &66.57&-\\ 

PBSL~\cite{shen2023pedestrian}&ACMMM'23 & RN50&BERT&65.32 &83.81 &89.26&-&-\\
BEAT\cite{ma2023beat}&ACMMM'23&RN101&BERT& 65.61& 83.45&89.54&-&-\\ 
LCR$^2$S~\cite{yan2023learning}&ACMMM'23&RN50&TextCNN&67.36& 84.19& 89.62& 59.24&\\
DCEL~\cite{li2023dcel}&ACMMM'23&CLIP-ViT& CLIP-Xformer &   75.02&  \textbf{90.89}&  \textbf{94.52}& -& -\\
UniPT~\cite{shao2023unified}&ICCV'23&CLIP-ViT& CLIP-Xformer &68.50& 84.67 &-&-&-\\ 
\color{gray!60} 
 RaSa~\cite{bai2023rasa}&\color{gray!60} IJCAI'23&\color{gray!60} ALBEFF&\color{gray!60} ALBEFF&\color{gray!60} 76.51& \color{gray!60} 90.29 &\color{gray!60} 94.25 &\color{gray!60} 69.38&\\
RaSa$_{\text{TCL}}$~\cite{bai2023rasa}&IJCAI'23&TCL&TCL&73.23& 89.20& 93.32& 66.43&-\\
TBPS~\cite{cao2023empirical}&Arxiv'23&CLIP-ViT& CLIP-Xformer&73.54& 88.19 &92.35& 65.38&-\\
\midrule
\textbf{Our RDE}& -& CLIP-ViT& CLIP-Xformer & \textbf{75.94} &{90.14}& {94.12} &\textbf{67.56}& \textbf{51.44} \\  

\bottomrule[1pt]
\end{tabular}}
\caption{Performance comparisons on the CUHK-PEDES dataset. The best results are in \textbf{bold}.} 
\label{tabcuhk}
\end{table*}

\begin{table}[h]
\centering

\resizebox{0.9\linewidth}{!}{
\begin{tabular}{l|ccccc}
\toprule[1pt] 
Methods    & R-1 & R-5 & R-10 & mAP   & mINP \\\midrule
Dual Path~\cite{zheng2020dual} & 38.99  & 59.44  & 68.41   & -     & -    \\
CMPM/C~\cite{zhang2018deep}    & 43.51  & 65.44  & 74.26   & -     & -    \\
ViTAA~\cite{wang2020vitaa}     & 50.98  & 68.79  & 75.78   & -     & -    \\
SSAN~\cite{ding2021semantically}      & 54.23  & 72.63  & 79.53   & -     & -    \\
IVT~\cite{shu2022see}       & 56.04  & 73.60  & 80.22   & -     & -    \\
ISANet~\cite{yan2022image}    & 57.73  & 75.42  & 81.72   & -     & -    \\
CFine~\cite{yan2022clip}     & 60.83  & 76.55  & 82.42   & -     & -    \\
IRRA~\cite{jiang2023cross}      & 63.46  & 80.25  & 85.82   & 38.06 & \textbf{7.93} \\
BiLMa~\cite{fujii2023bilma}  &
63.83& 80.15& 85.74& 38.26&-\\
PBSL~\cite{shen2023pedestrian}&57.84 &75.46& 82.15&-&-\\
BEAT\cite{ma2023beat}&58.25& 75.92 &81.96&-&-\\
LCR$^2$S~\cite{yan2023learning}&57.93 &76.08& 82.40 &38.21&-\\
DCEL~\cite{li2023dcel}&64.88& 81.34 &86.72&-&-\\
UniPT~\cite{shao2023unified}&60.09 &76.19&-&-&-\\
\color{gray!60} 
RaSa~\cite{bai2023rasa}&\color{gray!60}65.28&\color{gray!60} 80.40 &\color{gray!60}85.12 &\color{gray!60}41.29&-\\
RaSa$_{\text{TCL}}^*$~\cite{bai2023rasa}&63.29&79.36&84.36&39.23&-\\
TBPS~\cite{cao2023empirical}&65.05& 80.34 &85.47& 39.83&-\\
  
\midrule
\textbf{Our RDE} &\textbf{67.68}& \textbf{82.47}& \textbf{87.36}& \textbf{40.06} &{7.87} \\
\bottomrule[1pt]
\end{tabular}}
\caption{Performance comparisons on the ICFG-PEDES dataset. The best results are in \textbf{bold}. `*' indicates our reproducible results.}
\label{tabicfg}
\end{table}

\begin{table}[h]
\centering 
\resizebox{0.9\linewidth}{!}{
\begin{tabular}{l|ccccc}
\toprule[1pt] 
Methods & R-1& R-5 & R-10& mAP   & mINP  \\\midrule
DSSL~\cite{zhu2021dssl}   & 39.05& 62.60  & 73.95& -     & -     \\
SSAN~\cite{ding2021semantically}   & 43.50& 67.80  & 77.15& -     & -     \\
LBUL~\cite{wang2022look}   & 45.55& 68.20  & 77.85& -     & -     \\
IVT~\cite{shu2022see}    & 46.70& 70.00  & 78.80& -     & -     \\
CFine~\cite{yan2022clip}  & 50.55& 72.50  & 81.60& -     & -     \\
IRRA~\cite{jiang2023cross}  & 60.20 & 81.30  & 88.20 & 47.17 & 25.28 \\ 
BiLMA~\cite{fujii2023bilma}&61.20& 81.50& 88.80 &48.51&-\\
PBSL~\cite{shen2023pedestrian}&47.80& 71.40& 79.90&-&-\\
BEAT\cite{ma2023beat}&48.10 &73.10 &81.30&-&-\\
LCR$^2$S~\cite{yan2023learning}&54.95& 76.65& 84.70 &40.92&-\\
DCEL~\cite{li2023dcel}& 61.35& 83.95 &\textbf{90.45}&-&-\\
\color{gray!60} RaSa~\cite{bai2023rasa}&\color{gray!60}66.90&\color{gray!60}86.50  &\color{gray!60}91.35 &\color{gray!60}52.31&-\\
RaSa$_{\text{TCL}}^*$~\cite{bai2023rasa}&65.20&\textbf{84.05}&89.85&50.14&-\\
TBPS~\cite{cao2023empirical}&61.95& 83.55& 88.75& 48.26&-\\

\midrule
\textbf{Our RDE} &\textbf{65.35}& {83.95} &{89.90}& \textbf{50.88}& \textbf{28.08} \\
\bottomrule[1pt]
\end{tabular}}
\caption{Performance comparisons on the RSTPReid dataset. The best results are in \textbf{bold}. `*' indicates our reproducible results.}
\label{tabrstp}
\end{table}

\section{Study on the Selection Ratio \label{app5}} 
\begin{figure}[t]
    \centering
    \resizebox{0.6\linewidth}{!}{ 
    \includegraphics{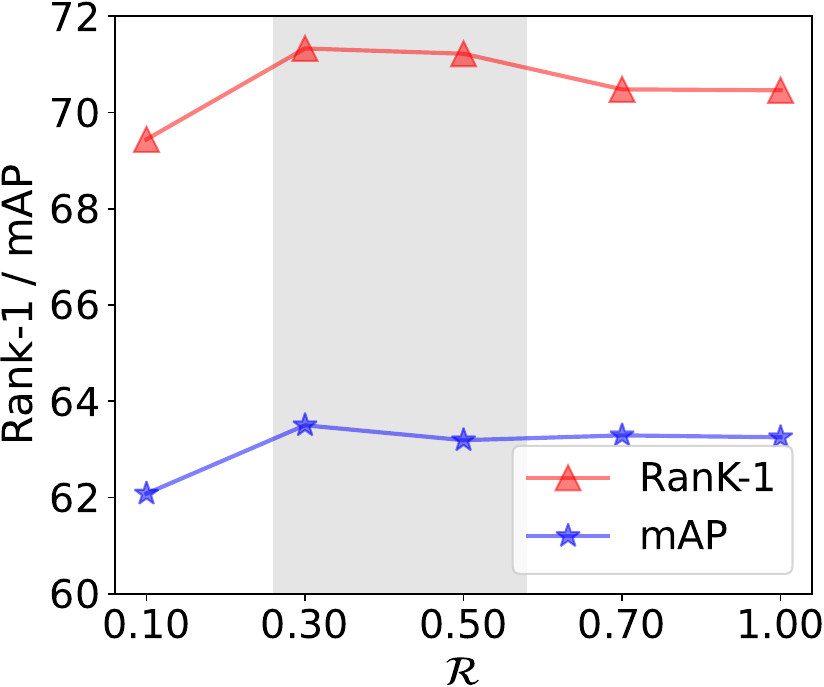}}
    \caption{Variation of performance with different $\mathcal{R}\in[0,1]$.}
    \label{fig_R_para}
\end{figure}

\Cref{fig_R_para} shows the variation of performance with different selection ratio $\mathcal{R}$. From the figure, one can see that too large or too small $\mathcal{R}$ will cause suboptimal performance. We think that a small $R$ will cause too much information loss and poor embedding presentations, while too large will focus on too many meaningless features. For this reason, we recommend $\mathcal{R}$ to be set between 0.3$\sim$0.5. Thus, $\mathcal{R}$ is set to 0.3 in all our experiments.
\begin{figure}[t]
    \centering
    \resizebox{\linewidth}{!}{ 
    \includegraphics{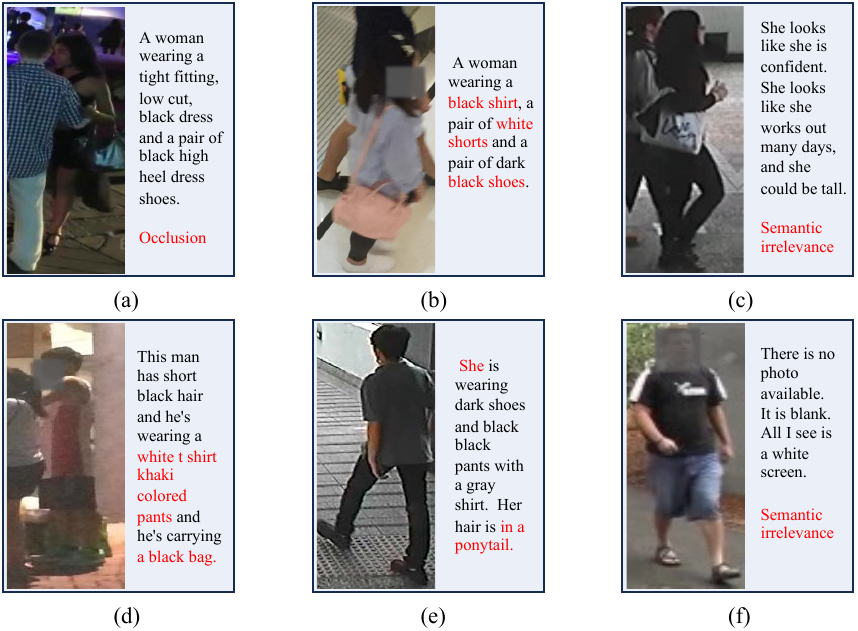}}
    \caption{The examples of noisy correspondence identified by CCD on the CUHK-PEDES dataset.}
    \label{fig0_}
\end{figure}
 \begin{figure}[t]
    \centering
    \resizebox{\linewidth}{!}{ 
    \includegraphics{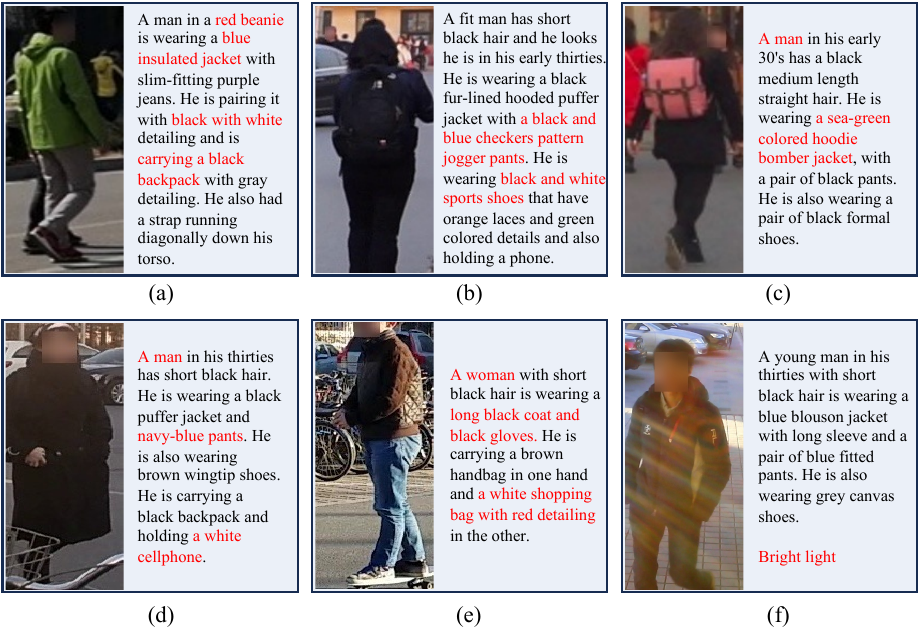}}
    \caption{The examples of noisy correspondence identified by CCD on the ICFG-PEDES dataset.}
    \label{fig1_}
\end{figure}
\begin{figure}[t]
    \centering
    \resizebox{\linewidth}{!}{ 
    \includegraphics{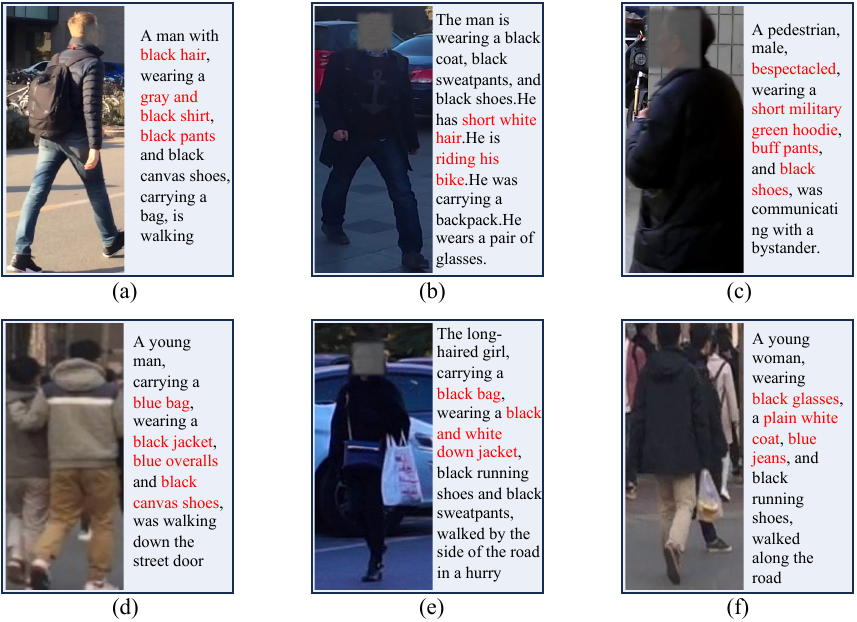}}
    \caption{The examples of noisy correspondence identified by CCD on the RSTPReid dataset.}
    \label{fig2_}
\end{figure}

\section{Ablation Study\label{app6}}
\subsection{Ablation analysis for TSE}

To verify the design rationality of TSE in our RDE, we conduct dedicated ablation experiments on TSE. The results are reported in~\Cref{tabTSE}. In the table, TSE$^\prime$ means that the token features encoded by CLIP are directly used for aggregation to obtain the embedding representations instead of conducting embedding transformation. Also, we show the impact of different pooling strategies on performance. From the results,  our standard version of TSE obtains the best performance, \ie, conducting the embedding transformation and using the max-pooling strategy to obtain the TSE representations.

\begin{table}[h]
\centering
\resizebox{\linewidth}{!}{
\begin{tabular}{c|l|ccccc}
\toprule[1pt] 
Methods &Pool& R-1& R-5 & R-10& mAP   & mINP  \\\midrule
TSE$^\prime$ &Avg.&67.22&84.96&90.03&60.22&43.84 \\
TSE$^\prime$ &TopK.&67.35&85.36&90.51&60.21&43.54 \\
TSE$^\prime$ &Max.&67.46&85.17&90.58&60.11&43.45\\
TSE &Avg.&67.43&85.19&90.50&60.42&43.97 \\
TSE &TopK.&68.27&86.03&90.79&60.95&44.37\\
TSE &Max.&\textbf{71.33}&\textbf{87.41}&\textbf{91.81}&\textbf{63.50}&\textbf{47.36}\\
\bottomrule[1pt]
\end{tabular}}
\caption{Performance comparisons with state-of-the-art methods on the RSTPReid dataset. 'Avg.', 'TopK.', and 'Max.' indicate the use of average-pooling, topK-pooling (K=10), and max-pooling strategies, respectively. }
\label{tabTSE}
\end{table}
\begin{table}[!h]
\centering
\setlength\tabcolsep{3.pt}
\resizebox{\linewidth}{!}{
\begin{tabular}{c|ccccc|ccccc}\toprule[1pt]
Noise&No.&$S^b$&$S^t$&CCD&Loss&R-1&R-5&R-10&mAP&mINP \\\midrule
\multirow{6}{*}{80\%}&\#1&\checkmark&\checkmark&\checkmark&TAL&\textbf{64.99}&\textbf{83.15}&\textbf{89.52}&\textbf{57.84}&\textbf{41.07} \\
&\#2&\checkmark&\checkmark&\checkmark&TRL&2.18&6.45&10.48&2.65&0.83\\
&\#3&\checkmark&\checkmark&\checkmark&TRL-S&51.62&74.53&82.21&46.15&30.12\\
&\#4&\checkmark&\checkmark&\checkmark&SDM&58.32&79.03&85.79&51.27&34.00 \\
&\#5&&\checkmark&\checkmark&TAL&63.56&82.59&88.84&56.69&39.71\\
&\#6&\checkmark&&\checkmark&TAL&61.70&81.61&87.95&55.11&38.34\\
&\#7&\checkmark&\checkmark&&TAL& 41.03&62.62&71.99&37.29&23.54 \\
\bottomrule[1pt]
    \end{tabular}}
    \caption{Ablation studies on the CHUK-PEDES dataset.}
    \label{tab_ab2}
\end{table}

\begin{figure*}[!h]
\centering
\subfloat[][CUHK-PEDES]{
\includegraphics[width=0.33\linewidth]{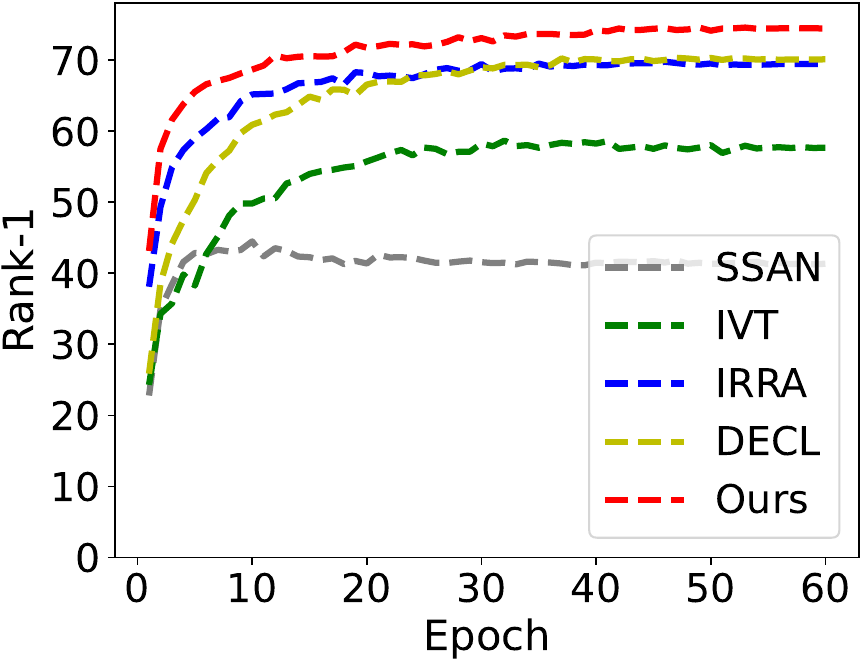}}
\subfloat[][ICFG-PEDES]{
\includegraphics[width=0.33\linewidth]{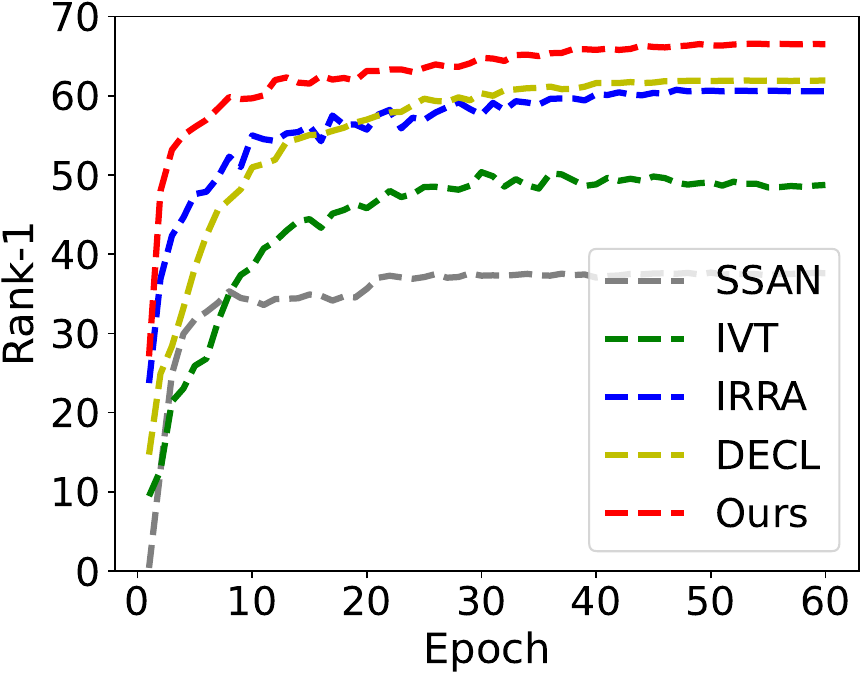}	}
\subfloat[][RSTPReid]{
\includegraphics[width=0.33\linewidth]{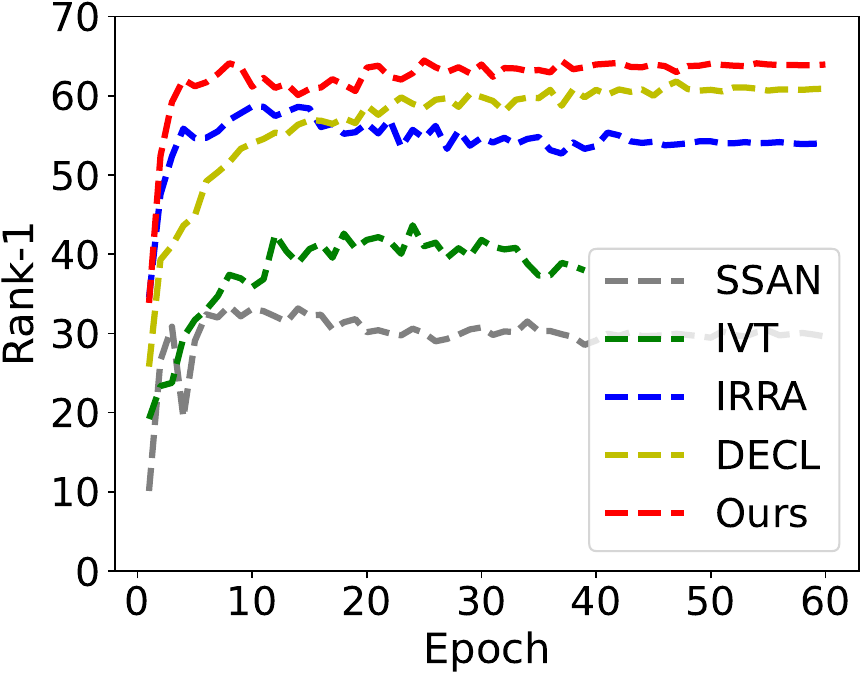}	}
\caption{Test performance (Rank-1) versus epochs on three datasets with 20\% noise.}
\label{fig0.2}
\end{figure*}
\begin{figure*}[!h]
\centering
\subfloat[][CUHK-PEDES]{
\includegraphics[width=0.33\linewidth]{img/CUHK-PEDES_acc_0.5_.pdf}}
\subfloat[][ICFG-PEDES]{
\includegraphics[width=0.33\linewidth]{img/ICFG-PEDES_acc_0.5_.pdf}}
\subfloat[][RSTPReid]{
\includegraphics[width=0.33\linewidth]{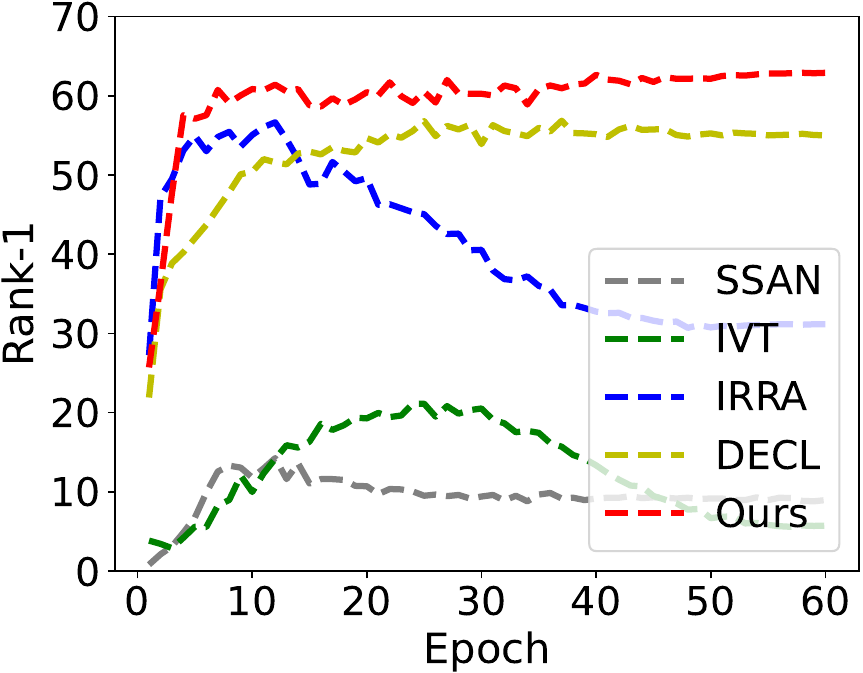}}
\caption{Test performance (Rank-1) versus epochs on three datasets with 50\% noise.}
\label{fig0.5}
\end{figure*}
\begin{figure*}[!h]
\centering
\subfloat[][CUHK-PEDES]{
\includegraphics[width=0.33\linewidth]{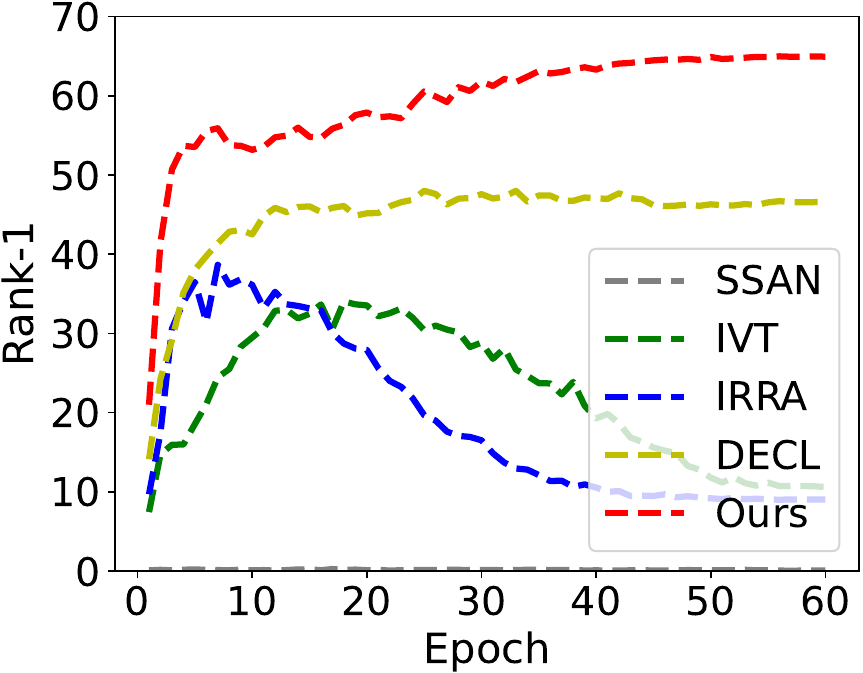}}
\subfloat[][ICFG-PEDES]{
\includegraphics[width=0.33\linewidth]{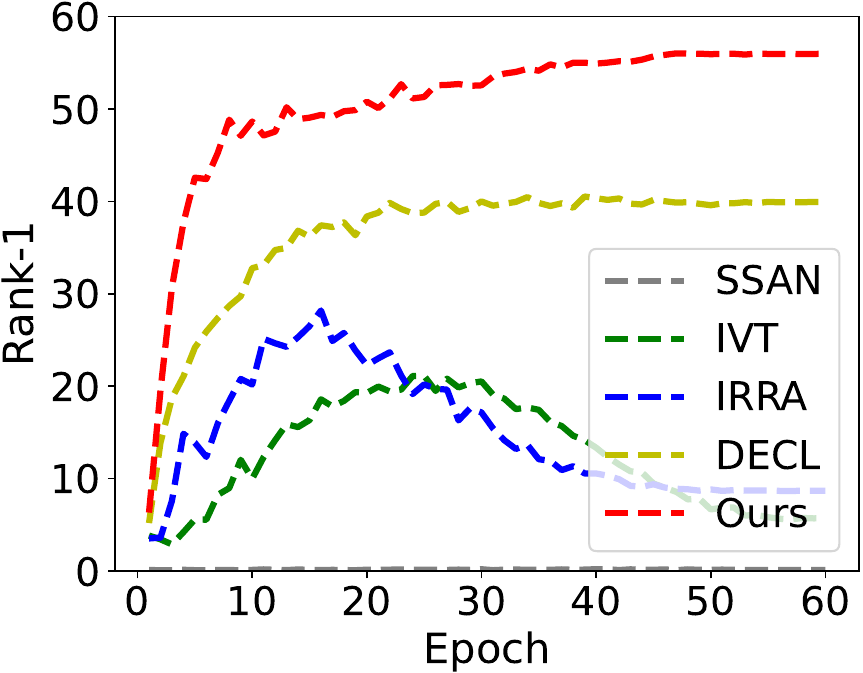}}
\subfloat[][RSTPReid]{
\includegraphics[width=0.33\linewidth]{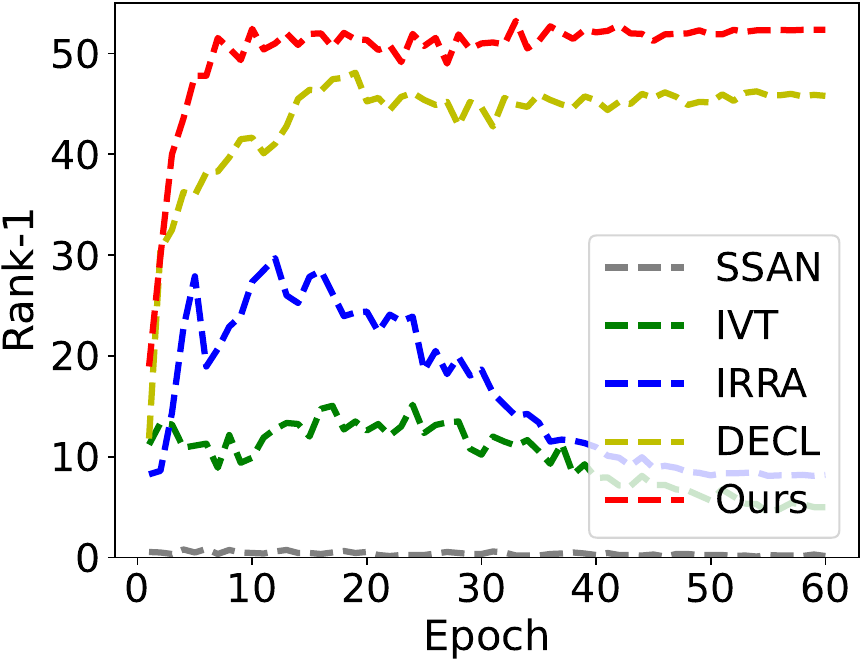}}
\caption{Test performance (Rank-1) versus epochs on three datasets with 80\% noise.}
\vspace{-0.3cm}
\label{fig0.8}
\end{figure*}

\subsection{Ablation study on High Noise}
In this appendix, we provide more ablation studies on the CUHK-PEDES dataset to investigate the effects and contributions of each proposed component in RDE. The experimental results are shown in \Cref{tab_ab2}. The observations and conclusions are consistent with those in the main text, which also demonstrate the effectiveness of our method.

\section{Case Study\label{app7}}
In this section, we show a large number of real examples of noisy pairs in three public datasets without synthetic NCs in~\Cref{fig0,fig1,fig2}, which are identified by CCD. Note that for privacy and security, the face areas of people in all images are \textbf{blurred}. From these examples, one can see that there are various reasons for noisy correspondences, \eg,  occlusion~(\eg,  \Cref{fig0_}(a,b)), lighting (\eg,  \Cref{fig1_}(f)), and inaccurate noisy text descriptions (\eg,  \Cref{fig0_}(c,f) and \Cref{fig2_}(a-f)). But all in all, these noisy pairs are real in these datasets and actually \textbf{break the implicit assumption} that all training image-text pairs are aligned correctly and perfectly at an instance level. Thus, we reveal the noisy correspondence problem in TIReID and propose a robust method, \ie, RDE, to particularly address it.

\begin{figure*}[!h]
    \centering
    \resizebox{\linewidth}{!}{ 
    \includegraphics{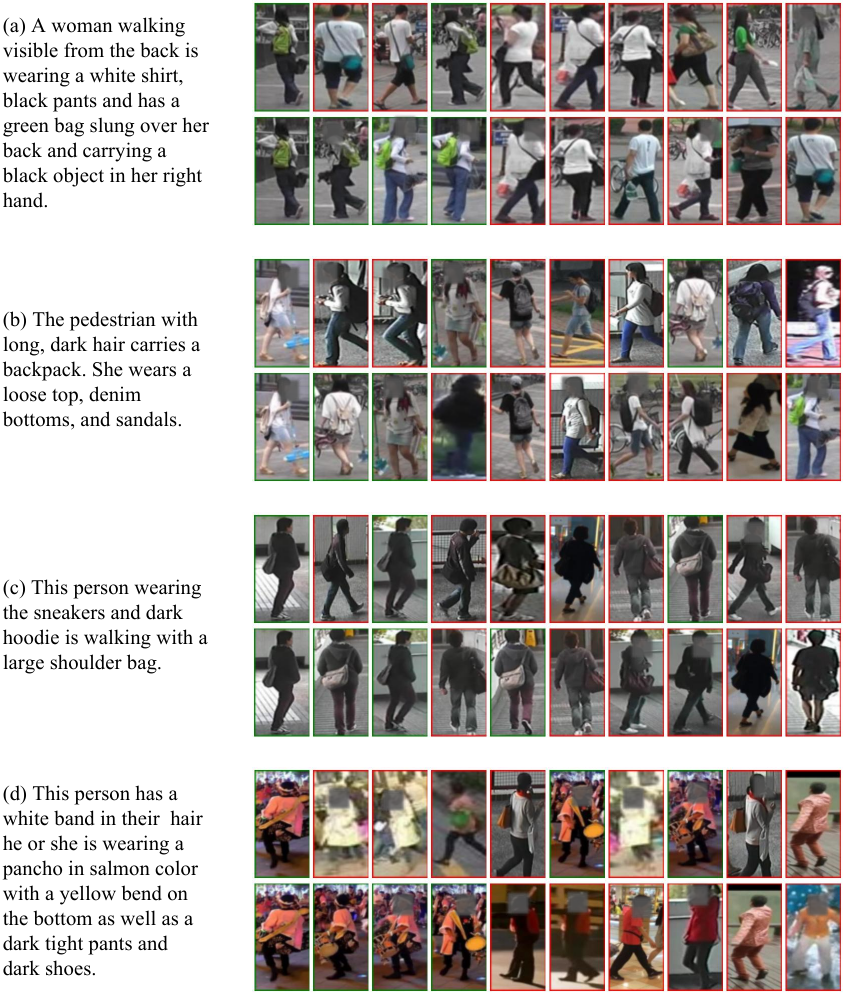}}
    \caption{Comparison of top-10 retrieved results on the CUHK-PEDES dataset between the baseline IRRA (the first row) and our RDE (the second row) for each text query. The matched and mismatched person images are marked with red and blue rectangles, respectively. All face areas of people in images are \textbf{blurred} for privacy and security.}
    \label{fig_egg}
\end{figure*}

\section{Robustness Study \label{app8}} 
For a comprehensive robustness analysis, we provide more performance curves versus epochs in~\Cref{fig0.2,fig0.5,fig0.8}. It can be seen from the~\Cref{fig0.2} that when the noise rate is 20\%, each baseline shows a certain degree of robustness, and there is no obvious performance degradation due to over-fitting noisy pairs. However, as the noise rate increases, the non-robust methods (SSAN, IVT, and IRRA) all show a curve that first rises and then falls. This tendency is caused by the memorization effect that DNNs tend to learn simple patterns before fitting noisy samples. Besides, we can also find that when the noise rate is 80\%, SSAN fails and other non-robust methods (IVT and IRRA) also have a serious performance drop. By contrast, thanks to the CCD and TAL, our RDE can learn accurate visual-semantic associations by obtaining confident clean training image-text pairs, which can effectively and directly prevent over-fitting noisy pairs, thus achieving robust cross-modal learning. From these figures, our method not only exhibits strong robustness but also achieves excellent re-identification performance.

\section{Qualitative Results\label{app9}}
To illustrate the advantages of our RDE, some retrieved examples
for TIReID are presented in~\Cref{fig_egg}. These results are obtained by testing the model trained on the CUHK-PEDES dataset with 20\% NCs. From the examples, one can see that our RDE obtains more accurate and reasonable re-identification results. Simultaneously, in some inaccurate results (\eg,  the results (b) and (d)) obtained by IRRA, we find that the visual information of the retrieved image often only matches part of the text query, which indicates that the model cannot learn complete alignment knowledge. We think the reason is that the NCs mislead the model of IRRA to focus on some wrong visual-semantic associations. In contrast, our RDE could filter out erroneous correspondences to learn reliable and accurate cross-modal knowledge, thus achieving high robustness and better results.

{
    \small
    \bibliographystyle{ieeenat_fullname}
    \bibliography{main}
}

\end{document}